\documentclass[10pt,twocolumn,letterpaper]{article}

\usepackage{cvpr}              % To produce the CAMERA-READY version
\usepackage[table]{xcolor}
\usepackage{stfloats}
\usepackage{caption}

\usepackage{amsmath}
\usepackage{amssymb}
\usepackage{bbm}
\usepackage{xcolor}
\usepackage{float}
\usepackage{multirow}

\usepackage{amsthm}
\usepackage{svg}
\usepackage[accsupp]{axessibility}  % Improves PDF readability for those with disabilities.

\newtheorem{theorem}{Theorem}
\newtheorem{repeat-theorem}{Theorem}
\newtheorem{lemma}{Lemma}
\newtheorem{repeat-lemma}{Lemma}
\newtheorem{corollary}{Corollary}
\newtheorem{repeat-corollary}{Corollary}
\newtheorem{assumption}{Assumption}

\newtheorem{proposition}{Proposition}
\newtheorem{repeat-proposition}{Proposition}

\makeatletter
\newif\ifcvpr@inappendix
\cvpr@inappendixfalse

\newcommand{\CVPRAppendixTOCMarker}{}

\newcommand{\MarkAppendixInTOC}{%
  \addtocontents{toc}{\protect\CVPRAppendixTOCMarker}%
}

\newcommand{\AppendixOnlyTOC}{%
  \begingroup
    \let\cvpr@orig@section\l@section
    \let\cvpr@orig@subsection\l@subsection
    \let\cvpr@orig@subsubsection\l@subsubsection
    \let\cvpr@orig@paragraph\l@paragraph
    \let\cvpr@orig@subparagraph\l@subparagraph
    \renewcommand{\l@section}[2]{\ifcvpr@inappendix \cvpr@orig@section{##1}{##2}\fi}
    \renewcommand{\l@subsection}[2]{\ifcvpr@inappendix \cvpr@orig@subsection{##1}{##2}\fi}
    \renewcommand{\l@subsubsection}[2]{\ifcvpr@inappendix \cvpr@orig@subsubsection{##1}{##2}\fi}
    \renewcommand{\l@paragraph}[2]{\ifcvpr@inappendix \cvpr@orig@paragraph{##1}{##2}\fi}
    \renewcommand{\l@subparagraph}[2]{\ifcvpr@inappendix \cvpr@orig@subparagraph{##1}{##2}\fi}
    \renewcommand{\CVPRAppendixTOCMarker}{\global\cvpr@inappendixtrue}
    \tableofcontents
  \endgroup
}
\makeatletter
\renewcommand{\@tocrmarg}{3em} % widen right margin before page number
\renewcommand{\@pnumwidth}{2em} % make page number column wider
\renewcommand{\l@section}{\@dottedtocline{1}{0em}{1.7em}} % increase left indent after "A"
\makeatother
\definecolor{cvprblue}{rgb}{0.21,0.49,0.74}

\newlength\savewidth\newcommand\shline{\noalign{\global\savewidth\arrayrulewidth\global\arrayrulewidth 1pt}\hline\noalign{\global\arrayrulewidth\savewidth}}
\newcommand{\tablestyle}[2]{\setlength{\tabcolsep}{#1}\renewcommand{\arraystretch}{#2}\centering\footnotesize}

\definecolor{demphcolor1}{gray}{.6}

\definecolor{eva02red}{RGB}{236,35,35}

\definecolor{02pink}{RGB}{240,178,188}

\definecolor{clipbaselinecolor}{gray}{.9}

\definecolor{defaultcolor}{HTML}{E8E2F7}

\definecolor{02skyblue}{RGB}{135,206,235}

\definecolor{02lightgray}{RGB}{180,180,180}
\newcommand{\rlightgray}{\rowcolor{02lightgray!25}}

\newcommand{\methodology}{BackSplit}

\newcolumntype{x}[1]{>{\centering\arraybackslash}p{#1pt}}
\newcolumntype{y}[1]{>{\raggedright\arraybackslash}p{#1pt}}
\newcolumntype{z}[1]{>{\raggedleft\arraybackslash}p{#1pt}}

\usepackage[pagebackref,breaklinks,colorlinks,allcolors=cvprblue]{hyperref}

\def\paperID{} % *** Enter the Paper ID here
\def\confName{CVPR}
\def\confYear{2026}

\title{BackSplit: The Importance of Sub-dividing the Background in Biomedical Lesion Segmentation}

\author{Rachit Saluja\textsuperscript{1,2,3}\thanks{\small Corresponding Author, \texttt{rs2492@cornell.edu}},
Asli Cihangir\textsuperscript{1},
Ruining Deng\textsuperscript{2,3},\\
Johannes C. Paetzold\textsuperscript{2,3},
Fengbei Liu\textsuperscript{2,3},
Mert R. Sabuncu\textsuperscript{1,2,3}
\and
\textsuperscript{1}Cornell University
\textsuperscript{2}Cornell Tech
\textsuperscript{3}Weill Cornell Medicine\\
}

\begin{document}
\maketitle
\begin{abstract}

Segmenting small lesions in medical images remains notoriously difficult. Most prior work tackles this challenge by either designing better architectures, loss functions, or data augmentation schemes; and collecting more labeled data. We take a different view, arguing that part of the problem lies in how the background is modeled. Common lesion segmentation collapses all non-lesion pixels into a single “background” class, ignoring the rich anatomical context in which lesions appear. In reality, the background is highly heterogeneous—composed of tissues, organs, and other structures that can now be labeled manually or inferred automatically using existing segmentation models.

In this paper, we argue that training with fine-grained labels that sub-divide the background class, which we call \textbf{\methodology}, is a simple yet powerful paradigm that can offer a significant performance boost without increasing inference costs. 
From an information theoretic standpoint, we prove that {\methodology} increases the expected Fisher Information relative to conventional binary training, leading to tighter asymptotic bounds and more stable optimization. With extensive experiments across multiple datasets and architectures, we empirically show that {\methodology} consistently boosts small-lesion segmentation performance, even when auxiliary labels are generated automatically using pretrained segmentation models. Additionally, we demonstrate that auxiliary labels derived from interactive segmentation frameworks exhibit the same beneficial effect, demonstrating its robustness, simplicity, and broad applicability. \href{https://rachitsaluja.github.io/backsplit/}{Project Page URL}.

\end{abstract}    
\vspace{-1.5em}
\section{Introduction}
\label{sec:intro}

Quantitative assessment of tumors and lesions represents one of the fundamental steps in clinical diagnosis and treatment planning. Despite substantial progress in medical image segmentation, particularly for anatomical structures such as organs — lesion segmentation remains a persistent challenge. Lesions are typically small, spatially sporadic, and underrepresented in dataset distributions, making them difficult to model reliably. These factors often lead to false positives and unstable predictions, ultimately limiting clinical deployment.

Prior research has primarily addressed this challenge through the development of improved network architectures, often tailored to specific tasks~\cite{ronneberger2015u, cciccek20163d, milletari2016v, chen2018encoder, wang2020deep, isensee2021nnu, chen2021transunet, hatamizadeh2022unetr, hatamizadeh2021swin}, or by designing specialized loss functions optimized for particular scenarios~\cite{kervadec2019boundary, abraham2019novel, zhu2019anatomynet, lin2017focal, luxtopograph, shit2021cldice}, and task-specific data augmentation techniques~\cite{yun2019cutmix, garcea2023data}. This focus is understandable, as medical image segmentation tasks vary considerably across imaging modalities, naturally motivating researchers to pursue modality- and task-specific architectural and optimization strategies.

In this work, we approach the problem from a different perspective by addressing a fundamental limitation shared across most existing architectures: the tendency to produce false positives and unstable lesion predictions due to their small size, irregular boundaries, and heterogeneous appearance.
In conventional setups, all non-lesion pixels are collapsed into a single “background” class, disregarding the rich anatomical context in which lesions occur. Yet, this background is far from homogeneous—it consists of diverse tissues, organs, and structures that can now be labeled manually or inferred automatically using modern segmentation models.

\begin{figure*}[t!]
  \centering
   \includegraphics[width=0.99\linewidth]{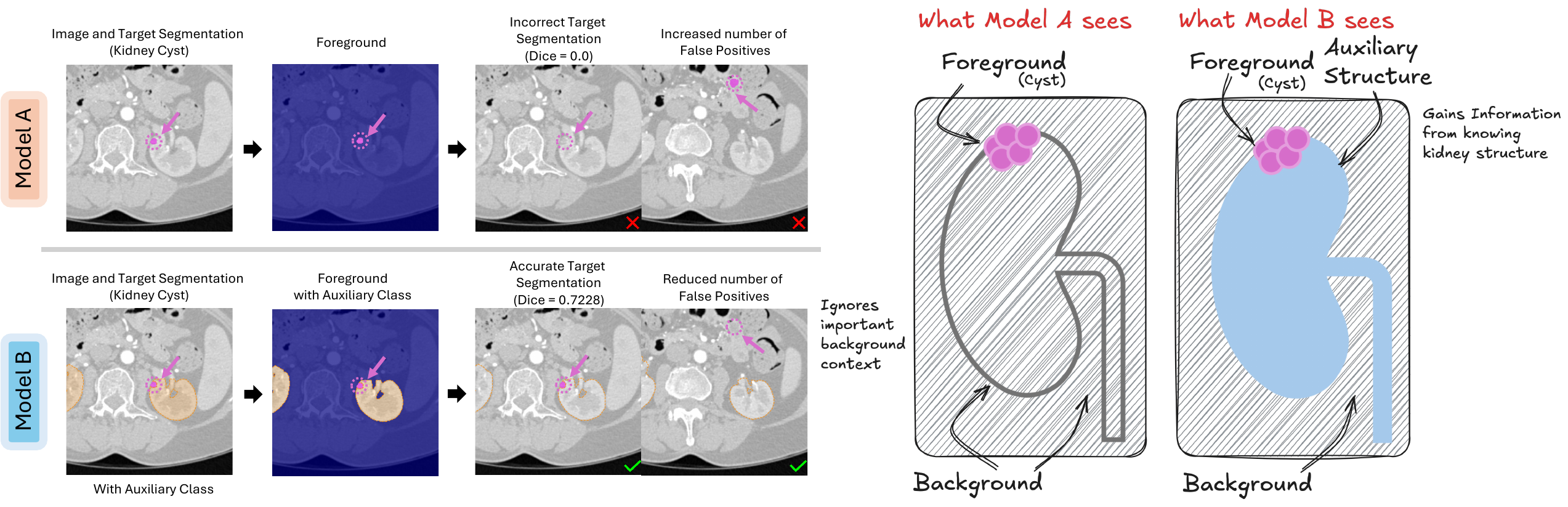}
   \caption{Comparison between conventional binary segmentation (Model A) and our BackSplit paradigm (Model B). Conventional lesion segmentation collapses all non-lesion regions into a single background, discarding the anatomical context and often producing false positives (yielding a Dice overlap score of 0.0). In contrast, BackSplit refines the background into semantically meaningful auxiliary structures (e.g., organ parenchyma) that are learned jointly with the lesion target. This structured background supervision enriches contextual understanding, yielding sharper lesion boundaries, fewer false detections (e.g., Dice = 0.72), and theoretically more stable predictions—consistent with higher expected Fisher Information and reduced estimator variance compared to conventional binary training.}
   \label{fig:teaser}
\end{figure*}

To address this limitation, we propose \textbf{\methodology} (as illustrated in \cref{fig:teaser}), a simple and intuitive training paradigm that incorporates structured background supervision. {\methodology} decomposes the background into multiple auxiliary classes and jointly optimizes them alongside the lesion target, thereby improving the supervision signal for the primary segmentation objective. This approach markedly reduces false positive detections and yields more accurate lesion boundaries. During inference, the model predicts the target class while implicitly leveraging the contextual knowledge learned from the auxiliary background classes, even though they are not explicitly used. 

In summary, {\methodology} redefines lesion segmentation by introducing structure into what is conventionally treated as a uniform background, improving both precision and generalization.

\noindent Our contributions are as follows: 
\begin{enumerate}
    \item We show that training with {\methodology} yields higher expected Fisher information compared to binary training (\cref{theorem:1}), leading to tighter asymptotic convergence bounds (\cref{corollary:1}).
    \item We validate the paradigm across five diverse datasets spanning multiple imaging modalities and anatomical regions, consistently improving performance metrics and reducing false positives across all architectures (\cref{tab:quantitative_comparison-1} and \cref{tab:quantitative_comparison-2}).
    \item We show that {\methodology} remains effective even when trained with automatically generated or noisy auxiliary segmentations,  making it more practical and easier to implement in real-world settings (\cref{tab:zoo_results} and \cref{tab:nni_results}).
    
\end{enumerate}

%% Code for adding Figures before the abstract
% \twocolumn[{
% \renewcommand\twocolumn[1][]{#1}%
% \maketitle
% \centering
% \vspace{-3em}
% \includegraphics[width=0.99\linewidth]{Figures/Figure_Teaser.pdf}
% \captionof{figure}{Conventional binary lesion segmentation (Model A) collapses all non-lesion regions into a single background, discarding the anatomical context in which lesions appear and often producing false positives. In contrast, {\methodology} (Model B) fine-grains the background into semantically meaningful auxiliary structures (e.g., organ parenchyma, vessels) learned jointly with the lesion target. This structured supervision enriches contextual understanding, yielding sharper boundaries, fewer false detections, and theoretically more stable predictions—reflecting higher Fisher information and lower estimator variance than conventional binary training.
% \vspace{1em}
% }
% \label{fig:teaser}
% }]
\section{Related Work}
\label{sec:Related Work}

\noindent \textbf{Binary vs Multi-Class Segmentation in Medical Imaging}. Medical image segmentation methods are commonly categorized into \textit{binary} and \textit{multi-class} paradigms, depending on how anatomical and pathological regions are delineated. In the binary setting, the task is formulated as distinguishing lesions from the background. This configuration remains prevalent in clinical pipelines, particularly for small or localized targets such as tumors~\cite{gros2021softseg}, nodules~\cite{hesamian2019atrous}, or infarcts~\cite{lecesne2023segmentation}. Such datasets typically provide lesion annotations without explicit organ or tissue masks, enforcing a binary formulation that limits contextual awareness of surrounding anatomy~\cite{dmitriev2019learning,fang2020multi,zhang2021dodnet,deng2023omni}.

Multi-head architectures have been proposed to approximate multi-class segmentation within a binary framework, where each head predicts an independent class mask~\cite{chen2019med3d,liu2024panoptic}. Although this setup allows training with incomplete labels, it neglects inter-class dependencies and often yields inconsistent anatomical boundaries.

In contrast, multi-class segmentation explicitly models multiple anatomical entities (e.g., organ parenchyma, lesions, and residual background) within a unified softmax formulation. By incorporating auxiliary contextual labels, the model learns to delineate lesion boundaries relative to their anatomical environment. Prior studies have shown that such contextual supervision enhances feature discriminability and improves generalization~\cite{huang2021deep,kuang2024weakly,yousaf2023multi}.

Our work advances this direction by introducing a theoretical justification and comprehensive empirical analysis of structured background supervision in lesion segmentation. For the first time, we provide a rigorous information-theoretic justification showing that multi-class contextual supervision yields higher expected Fisher Information, and therefore more stable estimators than conventional binary training. Complementing this theory, we perform extensive experiments across multiple architectures and datasets, including those with automatically and interactively generated auxiliary segmentations.

\noindent \textbf{Context‐aware and Auxiliary Supervision}. Contextual information plays a crucial role in distinguishing lesions from their surrounding anatomy, as many pathologies manifest through subtle appearance changes relative to neighboring tissues. Early studies leveraged organ masks or anatomical priors to stabilize lesion detection and constrain predictions within plausible regions~\cite{wang2019abdominal, zhang20213d, liu2021canet}. More recent approaches explicitly model cross‐class context through multi‐scale or anatomy‐guided reasoning, demonstrating that learning relationships among organs, tissues, and lesions substantially improves discriminability and generalization~\cite{deng2024prpseg, deng2024hats, jaus2024anatomy, sumesh, xu2025co}. These context‐aware frameworks employ strategies such as hierarchical decomposition, anatomy‐prompted supervision, and collaborative feature sharing across semantic levels to capture global anatomical coherence in addition to local lesion appearance. Multi‐task or shape‐aware regularization~\cite{he2020multi, li2025tri, stucki2023topologically, liu2023semi,liu2021shape} further enhance boundary precision by enforcing consistency between related anatomical structures. A similar intuition has been explored in~\cite{li2023context}, where an auxiliary network is introduced to predict background classes that enhance contextual understanding for target segmentation. In contrast, our work provides the theoretical foundation explaining why such an approach is effective.

While these context‐aware methods achieve impressive gains, they rely on additional architectural branches, explicit multi‐task losses, or handcrafted priors. In contrast, our {\methodology} paradigm is simply based on multi-class supervision: by decomposing the background into semantically meaningful auxiliary support structures, the model learns lesion–context relationships intrinsically through the label space rather than through network design. This supervision-level contextualization remains architecture‐agnostic, scales naturally to different backbones, and consistently enhances precision, boundary stability.

\noindent \textbf{Label Coarsening}. To develop our theoretical framework, we draw inspiration from prior work analyzing the effects of label coarsening across different supervision regimes. Existing studies primarily investigate methods that allow models trained on coarse labels to achieve performance comparable to those trained on fine-grained annotations, provided the available data is sufficiently informative~\cite{fotakis2021efficient}. Conversely, other approaches treat each example as its own class to capture fine-grained patterns and increase inter-class separability when the label set lacks sufficient granularity~\cite{xu2021weakly}. Notably,~\cite{cole2022label} demonstrates that selecting an appropriate level of label granularity during training can yield greater performance gains than choosing a more sophisticated model architecture.

These prior works largely lack a theoretical foundation to understand the role of label coarsening and how it affects statistical efficiency. In this paper, we present a theoretically grounded analysis that connects label granularity to the expected Fisher Information of the learning problem. Specifically, we show that sub-dividing the background into semantically meaningful auxiliary structures enriches the contextual representation and provably increases the expected Fisher Information, thereby improving the stability and accuracy of lesion segmentation.
\section{Methods}
\label{sec:Methods}

Let us formally demonstrate that multi-class training yields higher expected Fisher information than coarsened binary training, therefore providing more statistically efficient predictions for the target class.

\subsection{Setup and Notation}
\label{sec:Methods-Notations}

Let $(X,Y)$ be a pair of random data points where $X \in \mathcal{X}$ is the input, which in our case is an image (patch or pixel), and $Y \in \{1,\ldots,K\}$ is the class label at some pixel. We assume a joint distribution $p_{\theta}(X,Y) = p_{\theta}(Y \mid X)\,p(X)$ parameterized by $\theta$, which denotes the model parameters. 
\newline

\noindent \textbf{Coarsened Label Definition}. We focus on a particular foreground (target) class $c \in \{1,\ldots,K\}$, (e.g. a lesion of interest) and define the binary coarsened label indicating class $c$ as:

\vspace{-1em}
\begin{equation*}
    Z = \mathbbm{1}\{Y=c\} \in \{0,1\} 
\end{equation*}

\noindent where $\mathbbm{1}$ is the indicator function, so that $Z=1$ if $Y$ is the target class and $Z=0$ otherwise. 

In simple words: we are focusing on a single class $c$ and comparing two training scenarios: (i) the full \textit{multiclass} case with $K$ classes, vs. (ii) a \textit{collapsed binary} case distinguishing $c$ vs. not-$c$.
\newline

\noindent \textbf{Posterior Class Probabilities}. Let,  $\eta_k (x, \theta) = \mathbbm{P}_\theta (Y = k \mid X = x) $ and $\eta_c (x, \theta) = \mathbbm{P}_\theta (Z = 1 \mid X = x) $ denote the per-class and target-class posterior probabilities, respectively.
\newline

\noindent \textbf{Likelihoods and Score Functions}. For a dataset $\{(X_i, Y_i)\}_{i=i}^n$, the two corresponding log-likelihoods are

\vspace{-1em}
\begin{align*}
    \text{(multiclass)} \quad 
        & \ell_Y(\theta) = \sum_{i=1}^{n} \log p_\theta(Y_i \mid X_i) \\
    \text{(collapsed binary)} \quad 
        & \ell_Z(\theta) = \sum_{i=1}^{n} \log p_\theta(Z_i \mid X_i)
\end{align*}

\noindent where $p_\theta(Z=1 \mid x) = \eta_c(x;\theta)$ and $p_\theta(Z=0 \mid x) = 1-\eta_c(x;\theta) = \sum_{k\neq c} \eta_k(x;\theta)$. We define the per-sample score (gradient of the log-likelihood) and the observed information (negative Hessian) as

\vspace{-1em}
\begin{align*}
    s_Y (\theta) = \nabla_\theta \log p_\theta (Y \mid X) \\
    s_Z (\theta) = \nabla_\theta \log p_\theta (Z \mid X) \\
    I_Y (\theta) = - \nabla^2_\theta \log p_\theta (Y \mid X) \\
    I_Z (\theta) = - \nabla^2_\theta \log p_\theta (Z \mid X)
\end{align*}

\noindent \textbf{Expected Fisher Information}. Under standard regularity assumptions, the expected Fisher information matrices—equivalently the expectations of the observed information, are given by

\vspace{-1em}
\begin{align*}
    \mathcal{I}_Y (\theta) = \mathbb{E}_\theta[s_Y(\theta) s_Y(\theta)^T] = \mathbb{E}_\theta[I_Y(\theta)]\\
    \mathcal{I}_Z (\theta) = \mathbb{E}_\theta[s_Z(\theta) s_Z(\theta)^T] = \mathbb{E}_\theta[I_Z(\theta)]
\end{align*}

Conditioning on a fixed input $X=x$ gives:

\vspace{-1em}
\begin{align*}
    \mathcal{I}_Y (\theta \mid X=x) = \mathbb{E}_\theta[s_Y(\theta) s_Y(\theta)^T \mid X=x]
\end{align*}

\noindent where the expectation is taken over the posterior $p_\theta(Y\mid X)$. The unconditional Fisher Information integrates this quantity over $p(X)$.

\subsection{Expected Fisher information under label coarsening}

Having defined the score functions and expected Fisher Information for both the multiclass and coarsened (binary) formulations, we now formalize their relationship. The following lemma shows that the coarsened score is the conditional expectation (or orthogonal projection) of the full multiclass score given the observed variables.

\begin{lemma}[Score Projection~\cite{louis1982finding, oakes1999direct}]
Let $Z = g(Y)$ be a deterministic coarsening of the label $Y$.
Then, under regularity conditions ensuring that differentiation and summation interchange,
\begin{equation*}
    \mathbb{E}_\theta[s_Y(\theta)\mid Z,X] = s_Z(\theta).
\end{equation*}
\label{lemma:1}
\vspace{-1.5em}
\end{lemma}

\noindent [Proof in Supplementary \cref{sec:proofs-l1}]. Intuitively, \cref{lemma:1} (derived from the \emph{Missing Information Principle}~\cite{louis1982finding, oakes1999direct}) implies that $s_Z(\theta)$ is the best $L^2$ approximation of $s_Y(\theta)$ measurable with respect to $(Z,X)$. The coarsened score thus preserves only the gradient information recoverable from the coarsened labels while discarding variation across the collapsed classes.

Building on \cref{lemma:1}, we can now express the Fisher information of the complete labels $Y$ as the sum of the information contained in the coarsened labels $Z = g(Y)$ and a \emph{non-negative missing-information} term.
This decomposition formalizes how label coarsening (e.g., collapsing multiclass labels into a binary mask) can only reduce the available expected Fisher information.

\begin{theorem}[Label coarsening reduces expected Fisher information.]
For every $\theta\in\Theta$,
\begin{equation*}
  \mathcal{I}_Y(\theta)
  =\mathcal{I}_Z(\theta)
   +\mathbb{E}_\theta\!\big[\operatorname{Var}(s_Y(\theta)\mid Z,X)\big]
  \succeq\mathcal{I}_Z(\theta)
  \label{eq:fisher-decomposition}
\end{equation*}
and therefore $\mathcal{I}_Z(\theta)\preceq\mathcal{I}_Y(\theta)$ in the
Loewner (positive–semidefinite) order.
Equality holds iff $s_Y(\theta)$ is completely determined by $(Z,X)$,
i.e.\ no variation in the complete–data score remains once $Z$ is known.
\label{theorem:1}
\end{theorem}

\noindent [Proof in Supplementary \cref{sec:proofs-t1}] The coarsened Fisher information $\mathcal{I}_Z$ captures the portion of curvature in the log-likelihood that is recoverable from the observed variables $(Z,X)$,
while the conditional variance term quantifies the information lost
by collapsing fine-grained labels.

The equality condition in \cref{theorem:1} is particularly instructive. It states that if the complete score $s_Y (\theta)$ is fully determined by the coarsened variables $(Z,X)$ , then no Fisher information is lost. Intuitively, one can view $s_Y (\theta)$ as the gradient update that a sample $(X,Y)$ contributes during training. Suppose instead that only the coarsened pair $(X,Z)$ is observed. If every non-target label produces the same gradient direction given $X$ , then the gradient obtained from the coarsened label is identical to that from the complete label—no information gap arises, and $\mathcal{I}_Y (\theta) = \mathcal{I}_Z (\theta)$.

\begin{figure}[t]
  \centering
      \includegraphics[width=0.99\linewidth]{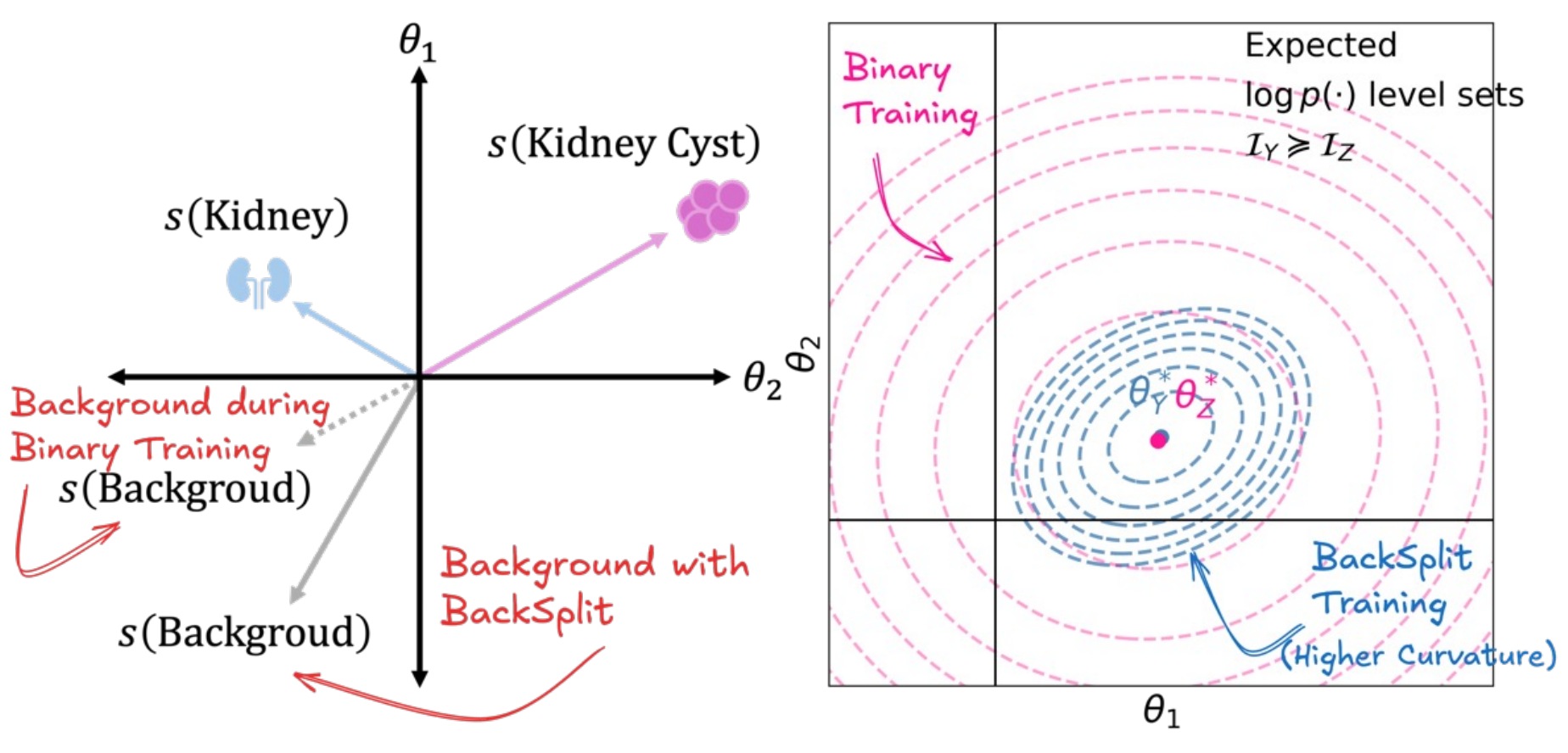}

   \caption{(Left) In binary training, the background gradient $s(\text{Background})$ conflates signals from multiple anatomical structures. Decomposing it into semantically distinct support structures (e.g., kidney) yields more disentangled and informative gradients. (Right) The corresponding log-likelihood level sets show that the decomposed formulation ($\mathcal{I}_Y$, blue) has sharper curvature and higher Fisher Information than the collapsed binary case ($\mathcal{I}_Z$, pink), leading to tighter and more stable parameter estimates.}
   \label{fig:intuition}
\vspace{-1.5em}
\end{figure}

In contrast, when different non-target classes induce different gradient directions (for instance, “organ A’’ and “organ B’’ yield distinct updates even though both are “not lesion’’), collapsing these labels to a single “background’’ class effectively averages those gradients. This averaging removes curvature directions in parameter space, reducing the total expected Fisher information (as shown in \cref{fig:intuition}). Consequently, for any input $X$ with non-zero target probability $p_\theta (Z=1 \mid X)$, if the non-target components of the multiclass gradient vary across classes, the collapsed binary model necessarily loses information. In other words, whenever non-target labels carry discriminative structure relevant to the target, the inequality in Theorem 1 becomes strict.

\subsection{Implication for Max Likelihood Estimator (MLE)}
\label{section:MLE}

We now analyze the implications of \cref{theorem:1} for the maximum-likelihood estimators (MLEs) trained with multiclass and coarsened (binary) labels.

\begin{corollary}[Asymptotic Efficiency of the Multiclass MLE.]
Under standard regularity and correct-specification assumptions at the true parameter $\theta^*$, the multiclass MLE is (weakly) more statistically efficient than the collapsed-binary MLE for any smooth functional of $\theta$. Equivalently, for any differentiable quantity $g(\theta)$ of interest, the asymptotic estimation variance of $g(\hat{\theta}_Y)$ is no greater than $g(\hat{\theta}_Z)$.
\label{corollary:1}
\end{corollary}

\noindent [Proof in Supplementary \cref{sec:proofs-c1}] At the optimum $\theta^*$, both estimators satisfy the classical MLE central-limit theorem:

\vspace{-1em}
\begin{align*}
        \sqrt{n} (\hat{\theta_Y} - \theta^*) \xrightarrow{d} \mathcal{N}(0,{\mathcal{I}_Y(\theta^*)}^{-1}) \\
        \sqrt{n} (\hat{\theta_Z} - \theta^*) \xrightarrow{d} \mathcal{N}(0,{\mathcal{I}_Z(\theta^*)}^{-1}) 
\end{align*}

Since \cref{theorem:1}, establishes the expected Fisher information ordering, $\mathcal{I}_Y(\theta)\succeq\mathcal{I}_Z(\theta)$, matrix inversion reverses this order for positive-definite matrices, and we obtain ${\mathcal{I}_Y(\theta)}^{-1}\preceq{\mathcal{I}_Z(\theta)}^{-1}$. Thus, the multiclass estimator has (weakly) smaller asymptotic covariance in parameter space.

The results thus far pertain to the quality of the parameter (weight) estimates. To relate this to the quality of model predictions, we employ the Delta method, which enables us to approximate the variance of prediction functions derived from these parameter estimates. Consider any smooth function $g(\theta)$ - for instance, the target-class posterior $\tau (\theta) = \eta_c(x_0;\theta)$, where $x_0$ denotes an arbitrary fixed input. By the delta method, 

\vspace{-1em}
\begin{align*}
        \sqrt{n} (g(\hat{\theta}) - g(\theta_0)) \xrightarrow{d} \mathcal{N}(0,G\Sigma G^T) 
\end{align*}

\noindent where $G = \nabla_\theta g(\theta_0)$, $\theta_0$ is a parameter with non-singular Fisher information, and $\Sigma$ is the asymptotic covariance of the corresponding MLE. As  $\Sigma_Y ={\mathcal{I}_Y(\theta)}^{-1} \preceq \Sigma_Z = {\mathcal{I}_Z(\theta)}^{-1}$, the variance of any differentiable prediction such as $p_\theta (Y=c \mid x_0)$ is strictly smaller (or equal) under multi-class training. 

% \vspace{-1em}
% \begin{multline*}
% \mathrm{Var}\!\left(\sqrt{n}\,(\tau(\hat{\theta}_Y) - \tau(\theta^*))\right)
% \le \mathrm{Var}\!\left(\sqrt{n}\,(\tau(\hat{\theta}_Z) - \tau(\theta^*))\right)
% \end{multline*}

\vspace{-1em}
\begin{multline*}
\mathrm{Var}\!\left(\sqrt{n}\,(\tau(\hat{\theta}_Y) - \tau(\theta^*))\right) \\
\le \mathrm{Var}\!\left(\sqrt{n}\,(\tau(\hat{\theta}_Z) - \tau(\theta^*))\right)
\end{multline*}

This formalizes the notion that multiclass training makes more efficient use of data. Both estimators are consistent and improve as $n \rightarrow \infty$ but the multiclass model converges faster and its confidence regions and prediction variances shrink more rapidly. Only when the equality condition of \cref{theorem:1} holds (the coarsened score fully determined by $(Z,X)$ do the two MLEs achieve identical efficiency. In all other cases, where non-target classes induce distinct gradient directions, the collapsed-binary estimator discards curvature information and remains asymptotically less precise.

\subsection{Special case: Softmax}

We now instantiate \cref{theorem:1} for the softmax layer which is commonly used for multi-class segmentation. This yields a closed-form expression for the information matrices in both the multiclass and the coarsened binary formulations.

\begin{proposition}[Softmax Expected Fisher Information Decomposition.]
Consider a softmax model with logits $f(x; \theta) \in \mathbb{R}^K$ and class probabilities $\eta = \text{softmax(f)}$. Let $J_f(x;\theta) = \partial f(x;\theta)/\partial \theta \in \mathcal{R}^{K \times p}$ denote the Jacobian of the logits with respect to the model parameters and let $e_c\in\mathbb R^{K}$ denote the $c$-th standard basis (one-hot) vector. Then for any fixed input $X=x$:
\begin{equation*}
  \mathcal{I}_Y(\theta \mid X=x)
  =J_f^T(\text{Diag}(\eta) - \eta \eta^T)J_f
\end{equation*}
is the conditional expected Fisher information for the multiclass likelihood, and
\begin{equation*}
  \mathcal{I}_Z (\theta \mid X=x) = J_f^T \left(\frac{\eta_c}{(1-\eta_c)} (e_c - \eta) {(e_c - \eta)}^T \right) J_f 
\end{equation*}
is the corresponding quantity for the collapsed-binary label. Hence, 
\begin{multline*}
\mathcal{I}_Y(\theta \mid X=x) = \mathcal{I}_Z(\theta \mid X=x) + \\
(1-\eta_c) J_f^T (\text{Diag}(\pi) - \pi \pi^T)J_f \succeq  \mathcal{I}_Z(\theta \mid X=x) 
\end{multline*}
where $\pi_k = \eta_k / (1- \eta_c)$ for $k \neq c$ and $\pi_c =0$.
\label{proposition:1}
\end{proposition}

\noindent [Proof in Supplementary \cref{sec:proofs-p1}] \cref{proposition:1} makes the information gap explicit for the softmax parameterization. The first term captures the curvature along the target-vs-rest direction $(e_c - \eta)$, identical to the expected Fisher information of the corresponding Bernoulli model. While the second term accounts for curvature within the non-target subspace of the probability simplex. Whenever at least two non-target classes have non-zero probability and the model parameters influence their relative proportions, this additional term is non-zero, yielding a strict inequality $\mathcal{I}_Y(\theta) \succ I_Z (\theta)$. Equality occurs only in degenerate cases (e.g., $K=2$ or a deterministic non-target distribution).

Geometrically, $\text{Diag}(\eta) - \eta \eta^T$ forms an ellipsoid of curvatures spanning the $K-1$ tangent directions of the simplex, while the collapsed term retains only the single direction $e_c - \eta$. Thus, label coarsening projects out all curvature within the non-target face, explaining the empirical loss of statistical efficiency observed in \cref{section:MLE}.

The softmax parameterization is particularly relevant to medical image segmentation, where most architectures employ a softmax output layer. Under this model, the Fisher decomposition exposes the information gap geometrically in logit space and directly relates it to practical scenarios such as lesion–background training. Collapsing non-target tissues into a single background class eliminates curvature directions that differentiate organs, thereby reducing Fisher information and impeding convergence.
\section{Experiments}
\label{sec:Experiments}

\begin{table*}[t!]
\centering
\tablestyle{4.2pt}{1.2}
    \begin{tabular}{r|cccc|cccc|cccc}
    & \multicolumn{4}{c|}{\textbf{U-Net}~\cite{isensee2021nnu, ronneberger2015u}} & \multicolumn{4}{c|}{\textbf{ResEncU-Net}~\cite{isensee2024nnu}} & \multicolumn{4}{c}{\textbf{SegResNet}~\cite{myronenko20183d}} \\
    \textbf{Task} &  \textbf{\#Params} & \textbf{Dice} $\uparrow$ & \textbf{HD-95} $\downarrow$ & \textbf{NSD} $\uparrow$ & \textbf{\#Params} & \textbf{Dice} $\uparrow$ & \textbf{HD-95} $\downarrow$ & \textbf{NSD} $\uparrow$ & \textbf{\#Params} & \textbf{Dice} $\uparrow$ & \textbf{HD-95} $\downarrow$ & \textbf{NSD} $\uparrow$\\
    \shline
    \shline
    \multicolumn{13}{c}{(a) Dataset: KiTS23 with Target Class = \{Cyst\}. Support Structures = \{Kidney, Tumors\}} \\
    \multicolumn{13}{c}{Total Sample Size (N = 489) | Imaging Modality = \{CT\}} \\
    \shline
    Cyst & 31.19M & 0.1787 &  428.4097 & 0.1695 & 102.35M & 0.2532 &  412.4152 & 0.2483 & 18.79M & 0.2914 &  374.4864 & 0.3554 \\ 
    \rlightgray
    Cyst (+\methodology) & 31.19M & \textbf{0.4573} & \textbf{267.2722} & \textbf{0.6004} & 102.35M & \textbf{0.4614} &  \textbf{259.6994} & \textbf{0.6100} & 18.79M & \textbf{0.4290} &  \textbf{287.6765} & \textbf{0.5580} \\ 
    \shline
    \shline
    \multicolumn{13}{c}{(b) Dataset: PANTHER-MR with Target Class = \{Tumor\}. Support Structures = \{Pancreas\}} \\
    \multicolumn{13}{c}{Total Sample Size (N = 92) | Imaging Modality = \{MRI\}} \\
    \shline
    Tumor & 30.70M & 0.4784 & 84.3689 & 0.2751 & 101.85M & 0.516 & 66.2637 & 0.2987 & 18.79M & 0.4744 & \textbf{59.0499} & 0.2623 \\ 
    \rlightgray
    Tumor (+\methodology) & 30.70M & \textbf{0.5251} & \textbf{54.2576} & \textbf{0.304} & 101.85M & \textbf{0.5165} & \textbf{70.5463} & \textbf{0.303} & 18.79M & \textbf{0.4995} & 73.1859 & \textbf{0.2778} \\ 
    \shline
    \shline
    \multicolumn{13}{c}{(c) Dataset: NSCLC-Radiomics with Target Class = \{GTV\}. Support Structures = \{Multiple Organs\}} \\
    \multicolumn{13}{c}{Total Sample Size (N = 415) | Imaging Modality = \{CT\}} \\
    \hline
    GTV & 30.70M & 0.4969 & 156.4725 & 0.3803 & 101.86M & 0.502 & 145.0108 & 0.3879 & 18.79M & 0.5061 & 142.8023 &\textbf{0.3921} \\ 
    \rlightgray
    GTV (+\methodology) & 30.70M & \textbf{0.5256} & \textbf{142.9498} & \textbf{0.4136} & 101.86M & \textbf{0.5386} & \textbf{125.9383} & \textbf{0.4268} & 18.79M & \textbf{0.5131} & \textbf{141.5307} & 0.3842 \\ 
    \shline
    \shline
    \end{tabular}
\caption{Quantitative comparison of segmentation performance across three architectures (U-Net~\cite{isensee2021nnu, ronneberger2015u}, ResEncU-Net~\cite{isensee2024nnu}, and SegResNet~\cite{myronenko20183d}) under standard and BackSplit training paradigms. Results are reported using 5-fold cross-validation on (a) KiTS (Cyst)~\cite{heller2023kits21}, (b) PANTHER-MR (Tumor)~\cite{betancourt_tarifa_2025_15192302}, and (c) NSCLC-Radiomics (GTV)~\cite{aerts2015data} datasets, with respective support structures indicated. BackSplit consistently improves Dice and NSD while reducing HD-95 across models and imaging modalities, without increasing model parameters (reported in millions).}
\label{tab:quantitative_comparison-1}
\vspace{-1.5em}
\end{table*}

We evaluate {\methodology} on five diverse and challenging datasets spanning CT, MRI, and PET modalities, and covering abdominal, thoracic, and whole-body regions. These datasets encompass a wide range of lesion segmentation tasks, architectures, and auxiliary structure labels, enabling a comprehensive assessment of robustness and generalizability even under label noise.

\subsection{Data}
The datasets collections comprise of:
\begin{enumerate}
    \item \textbf{KiTS23}~\cite{heller2023kits21}: This dataset is designed to facilitate the development of automated systems for semantic segmentation of kidneys and associated tumors and cysts. We specifically focus on kidney cysts as the target class. The dataset comprises 489 CT scans. 
    \item \textbf{PANTHER}~\cite{betancourt_tarifa_2025_15192302}: It comprises 92 annotated T1-weighted contrast-enhanced arterial phase MRI scans, each including manual annotations for both the pancreas and associated tumors.
    \item \textbf{NSCLC-Radiomics}~\cite{aerts2015data}: Dataset includes 415 CT scans from patients with non–small cell lung cancer (NSCLC), each containing manual segmentations of both thoracic organs and tumors. In our study, the gross tumor volume (GTV) serves as the target label.
    \item \textbf{AutoPET 3}~\cite{gatidis2022whole}:  This dataset consists of 1,611 paired CT and PET scans with corresponding lesion segmentations, encompassing a variety of cancer types. Our target labels are the tumors. 
    \item \textbf{MSWAL}~\cite{wu2025mswal}: It comprises 484 abdominal CT scans with multi-lesion annotations spanning seven classes, including stones, tumors, and cysts. All lesion categories are treated as target segmentations in our analysis.
\end{enumerate}

\subsection{{\methodology} is Agnostic to Network Architectures}

In our experiments, we trained two models for each setting:
(i) a binary baseline trained only on the target class as foreground, and
(ii) a {\methodology}-based model trained jointly on the target and its auxiliary support structures.

To assess architectural generality, we implemented {\methodology} across three widely adopted segmentation backbones that have achieved competitive performance in benchmark challenges:
(1) U-Net~\cite{ronneberger2015u},
(2) ResEnc U-Net~\cite{isensee2024nnu} implemented via the nnU-Net framework~\cite{isensee2021nnu}, and
(3) SegResNet~\cite{myronenko20183d}.

Since nnU-Net heuristically configures both architecture and hyperparameters, we ensured identical configurations between baseline and {\methodology} models for a fair comparison. All training followed the standard nnU-Net protocol, with detailed hyperparameters provided in the Supplementary Material.

\noindent \textbf{Results.} Our initial set of evaluations on the KiTS23, PANTHER, and NSCLC-Radiomics datasets were conducted with available ground-truth labels for the auxiliary classes, ensuring a controlled comparison. Each dataset was trained using five-fold cross-validation with all three architectures. Quantitative results are summarized in \cref{tab:quantitative_comparison-1}.
Across all datasets and architectures, {\methodology} consistently improves performance (measured with Dice, normalized surface distance, and the 95th-percentile Hausdorff distance). The auxiliary classes typically correspond to organs or tissues adjacent to the target structure (e.g., the pancreas for pancreatic tumors), demonstrating that structured background supervision improves lesion delineation without modifying the network architecture.

\subsection{{\methodology} works with model-derived auxiliary segmentations}

A key requirement for {\methodology} is the availability of auxiliary segmentations that can be used alongside the target class to enrich supervision. However, obtaining such annotations is often impractical as manual delineation of organ structures can require several hours per case. To overcome this limitation, we leverage existing organ segmentation models trained on large-scale datasets to automatically generate auxiliary labels through inference.

\begin{table}[t!]
\centering
\tablestyle{2.5pt}{1.2}
    \begin{tabular}{r|cccc}
    & \multicolumn{4}{c}{\textbf{U-Net}~\cite{isensee2021nnu, ronneberger2015u}} \\
    \textbf{Task} &  \textbf{\#Params} & \textbf{Dice} $\uparrow$ & \textbf{HD-95} $\downarrow$ & \textbf{NSD} $\uparrow$  \\
    \shline
    \shline
    \multicolumn{5}{c}{(a) Dataset: AutoPET with Target Class = \{Tumor\}} \\
    \multicolumn{5}{c}{Support Structures = \{Multiple Organs\}} \\
    \multicolumn{5}{c}{Total Sample Size (N = 1611) | Imaging Modality = \{PET, CT\}} \\
    \shline
    Tumor & 30.79M & 0.3881 &  637.6150 & 0.3478  \\ 
    \rlightgray
    Tumor (+\methodology) & 30.79M & \textbf{0.4435} & \textbf{594.8048} & \textbf{0.4128}\\ 
    \shline
    \shline
    \multicolumn{5}{c}{(b) Dataset: MSWAL with Target Class = \{Multiple Lesions\}} \\
    \multicolumn{5}{c}{Support Structures = \{Multiple Organs\}} \\
    \multicolumn{5}{c}{Total Sample Size (N = 484) | Imaging Modality = \{CT\}} \\
    \shline
    Gallstone & 30.79M &0.2665 & 418.338 & 0.6664\\ 
    \rlightgray
    Gallstone (+\methodology) & 30.79M & \textbf{0.3497} & \textbf{337.2602} & \textbf{0.7531}\\ 
    \hline
    Kidney Stone & 30.79M & \textbf{0.195} & \textbf{471.4653} & \textbf{0.6247} \\ 
    \rlightgray
    Kidney Stone (+\methodology) & 30.79M & 0.1639 & 484.3217 & 0.5401  \\ 
    \hline
    Liver Tumor & 30.79M & 0.2666 & 405.4476 & 0.3219  \\ 
    \rlightgray
    Liver Tumor (+\methodology) & 30.79M & \textbf{0.3312} & \textbf{338.2031} & \textbf{0.4683} \\ 
    \hline
    Kidney Tumor & 30.79M &0.1855 & 436.8838 & 0.6233  \\ 
    \rlightgray
    Kidney Tumor (+\methodology) & 30.79M & \textbf{0.1979} & \textbf{407.2282} & \textbf{0.6394}  \\ 
    \hline
    Pancreatic Cancer & 30.79M & 0.1836 & 448.5034 & 0.6018  \\ 
    \rlightgray
    Pancreatic Cancer (+\methodology) & 30.79M & \textbf{0.3228} & \textbf{308.2553} & \textbf{0.7746}  \\
    \hline
    Liver Cyst & 30.79M & 0.317 & 384.3436 & 0.4586 \\ 
    \rlightgray
    Liver Cyst (+\methodology) & 30.79M & \textbf{0.3571} & \textbf{341.4273} & \textbf{0.546} \\
    \hline
    Kidney Cyst & 30.79M & 0.4931 & 234.848 & 0.6072  \\ 
    \rlightgray
    Kidney Cyst (+\methodology) & 30.79M & \textbf{0.5104} & \textbf{213.0923} & \textbf{0.6311}  \\
    \shline
    All lesions mean & 30.79M & 0.2724 &  399.9756 & 0.5577 \\ 
    \rlightgray
    All lesions mean (+\methodology) & 30.79M & \textbf{0.3190} &   \textbf{347.1125} & \textbf{0.6218}  \\
    \shline
    \shline
    \end{tabular}
\caption{Quantitative results on (a) AutoPET and (b) MSWAL datasets using U-Net under standard and BackSplit training paradigms. For both datasets, support structures are automatically derived from a pre-trained organ segmentation model (U-Net trained on the AbdomenAtlas1.0Mini dataset). BackSplit yields consistent improvements in Dice and NSD, along with reductions in HD-95, across diverse lesion types and imaging modalities.}
\label{tab:quantitative_comparison-2}
\vspace{-1.5em}
\end{table}

\noindent \textbf{Results.} We implemented this strategy with AutoPET~\cite{gatidis2022whole} and MSWAL~\cite{wu2025mswal}, which provide only lesion annotations. Following the same setup as the previous experiment, we trained a U-Net on AbdomenAtlas1.0Mini~\cite{qu2023abdomenatlas} to obtain a pretrained organ segmentation model, then used it to automatically generate auxiliary masks for the CT scans of AutoPET and MSWAL. These model-derived organ labels were spatially consistent with the lesion regions and jointly used with the original lesion masks during training.

Despite the absence of manual auxiliary annotations, {\methodology} consistently improved Dice and surface-distance metrics, demonstrating robustness to automatically generated segmentations. As summarized in \cref{tab:quantitative_comparison-2}, performance gains were observed across all metrics. Notably, {\methodology} achieved higher accuracy on the multi-modal AutoPET dataset (CT + PET) and on MSWAL, where multiple lesion types were trained jointly with multiple organ structures. All experiments used a U-Net implemented via nnU-Net and were evaluated with five-fold cross-validation.

% \begin{table}[H]
% \centering
% \tablestyle{2.0pt}{1.2}
%     \begin{tabular}{r|ccc}
%     & \multicolumn{3}{c}{\textbf{U-Net}} \\
%     \textbf{Task} & \textbf{Dice} $\uparrow$ & \textbf{HD-95} $\downarrow$ & \textbf{NSD} $\uparrow$\\
%     \shline
%     \shline
%     \multicolumn{4}{c}{(a) PANTHER-MR} \\
%     \hline
%     Tumor & X & X & X \\
%     Tumor (+{\methodology}) & X & X & X \\
%     Tumor (+{\methodology} with VS-MR) & X & X & X \\
%     Tumor (+{\methodology} with TS-MR) & X & X & X \\
%     \shline
%     \shline
%     \multicolumn{4}{c}{(b) AutoPET} \\
%     \hline
%     Tumor & X & X & X \\
%     Tumor (+SS) & X & X & X \\
%     Tumor (+SS with VISTA3D) & X & X & X \\
%     Tumor (+SS with TotalSegmentator-CT) & X & X & X \\
%     \shline
%     \shline
%     \multicolumn{4}{c}{(c) MSWAL} \\
%     \hline
%     All Lesions & X & X & X \\
%     All Lesions (+SS) & X & X & X \\
%     All Lesions (+SS with VISTA3D) & X & X & X \\
%     All Lesions (+SS with TotalSegmentator-CT) & X & X & X \\
%     \shline
%     \shline
%     \end{tabular}
% \caption{\textbf{zero-shot retrieval performance (NEW.}}
% \label{tab:zoo_results}
% % \vspace{-1.5em}
% \end{table}

\begin{table}[H]
\centering
\tablestyle{1.5pt}{1.2}
\begin{tabular}{r|ccc|ccc}
& \multicolumn{3}{c|}{\textbf{(a) AutoPET}} 
& \multicolumn{3}{c}{\textbf{(b) MSWAL}} \\
\textbf{Method} 
& \textbf{Dice} $\uparrow$ & \textbf{HD-95} $\downarrow$ & \textbf{NSD} $\uparrow$
& \textbf{Dice} $\uparrow$ & \textbf{HD-95} $\downarrow$ & \textbf{NSD} $\uparrow$ \\
\shline
\shline
Regular Training
& 0.3921 & 677.2053 & 0.3415 
& 0.2518 & 415.4836 & 0.5335 \\
\rlightgray
\methodology 
& 0.4537 & 618.5841 & 0.4199 
& 0.2978 & 362.9077 & 0.6138 \\
\rlightgray
{\methodology} w/ TS~\cite{wasserthal2023totalsegmentator}
& 0.4456 & 642.5579 & 0.4036 
& 0.2843 & 374.2040 & 0.5932 \\
\rlightgray
{\methodology} w/ VS~\cite{graf2025vibesegmentator}
& 0.4314 & 682.6587 & 0.3761
& 0.2646 & 391.8890 & 0.5638 \\
\shline
\shline
\end{tabular}
\caption{Evaluation of {\methodology} on (a) AutoPET (tumors) and (b) MSWAL (all lesions) for a single fold using a U-Net backbone. Organ masks from TotalSegmentator (TS)~\cite{wasserthal2023totalsegmentator} and VIBE-Segmentator (VS)~\cite{graf2025vibesegmentator} were used as automatically extracted auxiliary structures. {\methodology} consistently improves Dice, HD-95, and NSD scores, demonstrating that even automatically inferred support segmentations yield measurable gains.}
\label{tab:zoo_results}
\vspace{-1.5em}
\end{table}

\subsection{{\methodology} using pre-trained large models}

{\methodology} can also leverage organ segmentations extracted from large pretrained models such as TotalSegmentator~\cite{wasserthal2023totalsegmentator,d2024totalsegmentator} and VIBE-Segmentator~\cite{graf2025vibesegmentator}, which support both CT and MR modalities. Using the same experimental setup as in the previous subsection, we generated auxiliary organ masks with these models and incorporated them during training to evaluate whether performance gains persist when auxiliary labels are obtained entirely through automated inference.

\noindent \textbf{Results.}
As summarized in \cref{tab:zoo_results}, {\methodology} continues to improve performance across Dice, HD-95, and NSD metrics for both AutoPET and MSWAL. This demonstrates that even simple, pretrained model inferences performed prior to training can yield a measurable and consistent performance gain.

\subsection{{\methodology} using Interactive Segmentation}

With recent advances in interactive segmentation foundation models, we further evaluate {\methodology} under realistic conditions where auxiliary segmentations are noisy or partially accurate. To simulate this scenario, we employ nnInteractive~\cite{isensee2025nninteractive}, a 3D interactive segmentation model. For each auxiliary structure, seven and ten random positive clicks are sampled from the ground-truth masks and used as seed inputs to nnInteractive. This procedure is applied to the KiTS23, NSCLC-Radiomics, and PANTHER datasets. For AutoPET and MSWAL, we follow the same protocol using the automatically generated auxiliary segmentations from our pretrained model as input to nnInteractive. The resulting pseudo-annotations are spatially aligned with the original lesion masks and used for joint training following the same setup as the previous experiments. This experiment evaluates the robustness of {\methodology} when trained with imperfect, interactively generated auxiliary supervision.

\noindent \textbf{Results.} As shown in the Supplementary Material, the auxiliary segmentations generated using nnInteractive exhibit modest accuracy. Nevertheless, when incorporated into U-Net training, we observe consistent performance improvements across all five datasets for both the 7-click and 10-click configurations (\cref{tab:nni_results}). These findings demonstrate that even highly noisy, interactively generated auxiliary cues can enhance target segmentation performance under the {\methodology} paradigm, highlighting its robustness to imperfect supervision.

\begin{table*}[t]
\centering
\tablestyle{1.8pt}{1.2}
\begin{tabular}{r|ccc|ccc|ccc|ccc|ccc}
& \multicolumn{3}{c|}{\textbf{(a) KiTS}} 
& \multicolumn{3}{c|}{\textbf{(b) PANTHER-MR}} 
& \multicolumn{3}{c|}{\textbf{(c) NSCLC-Rad}} 
& \multicolumn{3}{c|}{\textbf{(d) AutoPET}} 
& \multicolumn{3}{c}{\textbf{(e) MSWAL}} \\
\textbf{Method} 
& \textbf{Dice} $\uparrow$ & \textbf{HD-95} $\downarrow$ & \textbf{NSD} $\uparrow$
& \textbf{Dice} $\uparrow$ & \textbf{HD-95} $\downarrow$ & \textbf{NSD} $\uparrow$
& \textbf{Dice} $\uparrow$ & \textbf{HD-95} $\downarrow$ & \textbf{NSD} $\uparrow$
& \textbf{Dice} $\uparrow$ & \textbf{HD-95} $\downarrow$ & \textbf{NSD} $\uparrow$
& \textbf{Dice} $\uparrow$ & \textbf{HD-95} $\downarrow$ & \textbf{NSD} $\uparrow$ \\
\shline
\shline
Regular Training
& 0.2033 & 425.3273 & 0.1906 
& 0.3828 & 157.3470 & 0.2320 
& 0.5279 & 140.2698 & 0.4049 
& 0.3921 & 677.2053 & 0.3415 
& 0.2518 & 415.4836 & 0.5335 \\
\rlightgray
\methodology 
& 0.5297 & 249.5371 & 0.6703 
& 0.4906 & 54.2010 & 0.2796 
& 0.5862 & 124.2557 & 0.4533 
& 0.4537 & 618.5841 & 0.4199 
& 0.2978 & 362.9077 & 0.6138 \\
\rlightgray
{\methodology} with \textcircled{+}7 
& 0.4919 & 267.1777 & 0.6303 
& 0.5079 & 74.6472 & 0.2995 
& 0.5869 & 112.1321 & 0.4559 
& 0.4562 & 620.5639 & 0.4217 
& 0.2714 & 385.2560 & 0.5748 \\
\rlightgray
{\methodology} with \textcircled{+}10 
& 0.4921 & 276.7345 & 0.6195 
& 0.4866 & 74.6971 & 0.2799 
& 0.5746 & 113.6748 & 0.4470 
& 0.4649 & 602.8102 & 0.4422 
& 0.2805 & 378.3928 & 0.5860 \\
\shline
\shline
\end{tabular}
\caption{Single-fold evaluation of {\methodology} using nnInteractive-derived support structures. Results are reported for (a) KiTS23, (b) PANTHER-MR, (c) NSCLC-Rad, (d) AutoPET, and (e) MSWAL. Even when auxiliary structures are generated from noisy nnInteractive segmentations with 7- and 10-click simulated user inputs (see Supplementary), {\methodology} consistently outperforms standard training.}
\label{tab:nni_results}
\vspace{-1.5em}
\end{table*}

\begin{figure}[t]
  \centering
      \includegraphics[width=0.99\linewidth]{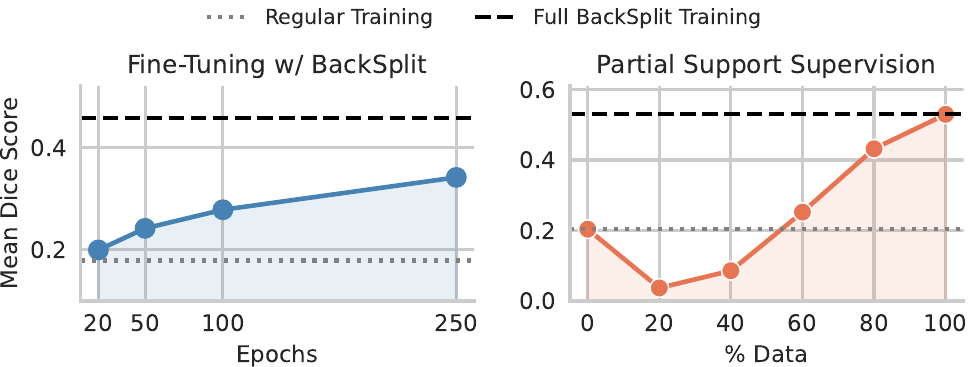}

   \caption{\textbf{(Left)} Fine-tuning a pretrained binary model with auxiliary structures steadily improves performance across epochs. \textbf{(Right)} Under partial support supervision, limited auxiliary data initially reduce performance but later yield consistent gains as more auxiliary structures are added, approaching full {\methodology} performance. Evaluated on KiTS23 (Target is Cyst).}
   \label{fig:ft_datamix}
   \vspace{-1.5em}
\end{figure}

\subsection{Behavior of {\methodology} under Fine-Tuning and Partial Supervision}

To explore how researchers can readily adopt the BackSplit paradigm, we conduct two complementary experiments using KiTS23 with cyst segmentation as the target task. First, we fine-tune a pretrained U-Net by incorporating auxiliary support structures. Second, we analyze the effect of training from scratch when only a subset of samples includes auxiliary annotations. Together, these experiments illustrate how existing models and datasets can be easily adapted to achieve improved performance with minimal modification.

\noindent \textbf{Results.} As shown in \cref{fig:ft_datamix} (Left), fine-tuning a pretrained binary model with auxiliary support structures yields immediate performance improvements in Dice score, even after just 50 epochs of training. The performance continues to improve with additional training, nearly doubling by 250 epochs compared to the original binary model. In \cref{fig:ft_datamix} (Right), we observe an interesting trend: when only a small fraction of auxiliary labels is included, models trained from scratch initially underperform, likely due to confusion between target and auxiliary structures. However, as the proportion of auxiliary labels increases linearly, the model gradually recovers and begins to exhibit the expected performance improvements characteristic of {\methodology}.

\section{Conclusion}
\label{sec:Conclusion}

We introduced {\methodology}, a simple and theoretically grounded training paradigm that improves lesion segmentation by decomposing the background into semantically meaningful auxiliary structures. From an information-theoretic standpoint, we showed that structured supervision increases the expected Fisher Information over conventional binary training, leading to more stable estimators. Experiments across five datasets, spanning multiple imaging modalities and model architectures, confirm {\methodology}’s effectiveness and generality.

We further validate {\methodology} under realistic conditions using automatically derived and noisy auxiliary segmentations from large pretrained models and interactive frameworks. Across all settings, BackSplit maintains consistent performance gains. Although our theoretical analysis assumes a large-sample regime—where finer label granularity increases Fisher curvature—we acknowledge that, in small-data scenarios, this could amplify sampling noise and risk overfitting. Empirically, however, we did not observe such effects within typical medical segmentation dataset sizes.

In future work, researchers may explore which specific auxiliary structures contribute most to performance improvements. Another promising direction is the use of language-driven or proxy representations of anatomical context (e.g., textual cues such as “liver” or “kidney”) to provide structural information without requiring explicit segmentations. Overall, BackSplit offers a strong and extensible paradigm that enhances lesion segmentation and supports clinical decision-making. 

{
    \small
    \bibliographystyle{ieeenat_fullname}
    \bibliography{main}
}

% \clearpage
% \setcounter{page}{1}
\maketitlesupplementary

\appendix
\renewcommand{\thesection}{\Alph{section}}

\MarkAppendixInTOC  
\setcounter{tocdepth}{2}
\AppendixOnlyTOC

\section{Full Proofs and Theoretical Insights}
\label{sec:proofs}

Our goal is to prove that using a full $K$-class training regime we yield a greater Fisher information than training on a coarsened/collapsed binary label. This in turn leads to more statistically efficient predictions for the target class.

\subsection{Setup and Notation}
\label{sec:proofs-notation}

The notations used to prove the following sections are used from \cref{sec:Methods-Notations}.

\begin{assumption}[Regularity and Optimization / Compute]
We work with a parametric family $\{p_\theta(Y\mid X) : \theta \in \Theta\}$,
where $\Theta \subset \mathbb{R}^p$ is an open set containing the true
parameter $\theta^\star$. We assume:
\begin{enumerate}
  \item \textbf{Model regularity.} For each fixed input $x$ the conditional
  density $p_\theta(Y\mid X=x)$ is correctly specified at $\theta^\star$ and
  is twice continuously differentiable in $\theta$ in a neighbourhood of
  $\theta^\star$. Differentiation and summation over $Y$ can be interchanged,
  and the Fisher information matrices
  \begin{multline*}
  I_Y(\theta) = \mathbb{E}_\theta\big[s_Y(\theta)s_Y(\theta)^\top\big], \\
  \qquad
  I_Z(\theta) = \mathbb{E}_\theta\big[s_Z(\theta)s_Z(\theta)^\top\big]
  \end{multline*}
  exist and are finite. At $\theta^\star$, $I_Y(\theta^\star)$ and
  $I_Z(\theta^\star)$ are non–singular.

  \item \textbf{Common architecture.} The multiclass and collapsed–binary
  models share the same network architecture and parameterization up to the
  final classification layer, so that both likelihoods are defined on the same
  parameter space $\Theta$.

  \item \textbf{Optimization / compute parity.} In practice, both models are
  trained with the same optimization algorithm and comparable compute
  budgets (e.g., same number of epochs, batch sizes, and learning–rate
  schedules), and we treat the resulting estimators as approximate
  maximum–likelihood estimators. In particular, we assume that training
  converges to stationary points that lie in a neighbourhood of the (local)
  maximizers of the corresponding log–likelihoods, so that the usual MLE
  asymptotics apply.
\end{enumerate}
\label{assum1}
\end{assumption}

Under Assumption 1, the standard MLE central limit theorem and delta method
results used in ~\cref{sec:Methods} and ~\cref{sec:proofs} hold for both the multiclass and
collapsed–binary estimators.

\subsection{Proof of Lemma 1}
\label{sec:proofs-l1}

\begin{repeat-lemma}[Score Projection~\cite{louis1982finding, oakes1999direct}]
Let $Z = g(Y)$ be a deterministic coarsening of the label $Y$.
Then, under regularity conditions ensuring that differentiation and summation interchange,
\begin{equation*}
    \mathbb{E}_\theta[s_Y(\theta)\mid Z,X] = s_Z(\theta).
\end{equation*}
\label{replemma:1}
\vspace{-1.5em}
\end{repeat-lemma}

\noindent \textbf{Proof:} Since $Z=g(Y)$, we can write:

\begin{align*}
\mathbb{E}[s_Y(\theta) \mid Z,X]
    &= \sum_{y:g(y) = Z} \nabla_\theta \log p_\theta (y \mid X)
       \frac{p_\theta (y \mid X)}{p_\theta (Z \mid X)} \\
    &= \frac{\sum_{y:g(y)=Z}\nabla_\theta p_\theta (y \mid X)}{p_\theta (Z \mid X)} \\
    &= \nabla_\theta \log p_\theta (Z \mid X) \\
    &= s_Z (\theta)
\end{align*}

\qed

\subsection{Proof of Theorem 1}
\label{sec:proofs-t1}

\begin{repeat-theorem}[Label coarsening reduces expected Fisher information.]
For every $\theta\in\Theta$,
\begin{equation*}
  \mathcal{I}_Y(\theta)
  =\mathcal{I}_Z(\theta)
   +\mathbb{E}_\theta\!\big[\operatorname{Var}(s_Y(\theta)\mid Z,X)\big]
  \succeq\mathcal{I}_Z(\theta)
  \label{eq:fisher-decomposition-rep}
\end{equation*}
and therefore $\mathcal{I}_Z(\theta)\preceq\mathcal{I}_Y(\theta)$ in the
Loewner (positive–semidefinite) order.
Equality holds iff $s_Y(\theta)$ is completely determined by $(Z,X)$,
i.e.\ no variation in the complete–data score remains once $Z$ is known.
\label{reptheorem:1}
\end{repeat-theorem}

\noindent \textbf{Proof:} From the law of total variance, for any random vector $U$ and any conditional variable $V$, 

\begin{align*}
    \mathbb{E} [UU^T] = \mathbb{E}\!\left[\mathbb{E}[U\mid V]\, {\mathbb{E}[U\mid V]}^T\right] + \mathbb{E}[\mathrm{Var}(U \mid V)]
\end{align*}

\noindent substituting $U = s_Y (\theta)$ and $V = (Z,X)$ we get.

\begin{multline*}
    \mathbb{E}_\theta [s_Y (\theta) {s_Y (\theta)}^T] = \\ \mathbb{E}_\theta  \!\left[\mathbb{E}_\theta[ s_Y (\theta) \mid (Z,X)]\, {\mathbb{E}_\theta[s_Y (\theta) \mid (Z,X)]}^T\right] \\ + \mathbb{E}_\theta[\mathrm{Var}(s_Y(\theta) \mid (Z,X))]
\end{multline*}

\noindent From \cref{lemma:1}, we get that $\mathbb{E}[s_Y(\theta) \mid Z,X] = s_Z (\theta)$.

\begin{multline*}
\mathbb{E}_\theta [s_Y (\theta) {s_Y (\theta)}^T] 
        = \mathbb{E}_\theta [s_Z(\theta){s_Z(\theta)}^T] \\ + \mathbb{E}_\theta[\mathrm{Var}(s_Y(\theta) \mid (Z,X))]     
\end{multline*}

\noindent From the definitions of expect Fisher Information (from \cref{sec:Methods-Notations}), we get:

\begin{align*}
        \mathcal{I}_Y (\theta)
        &= \mathcal{I}_Z (\theta) + \mathbb{E}_\theta[\mathrm{Var}(s_Y(\theta) \mid (Z,X))] 
\end{align*}
    
\noindent Additionally, we can define the missing information as 

\begin{align*}
        \mathcal{I}_{\text{missing}} =  \mathbb{E}_\theta[\mathrm{Var}(s_Y(\theta) \mid (Z,X))] 
\end{align*}

\qed

\subsection{Proof of Corollary 1}
\label{sec:proofs-c1}

\begin{repeat-corollary}[Asymptotic Efficiency of the Multiclass MLE.]
Under standard regularity and correct-specification assumptions at the true parameter $\theta^*$, the multiclass MLE is (weakly) more statistically efficient than the collapsed-binary MLE for any smooth functional of $\theta$. Equivalently, for any differentiable quantity $g(\theta)$ of interest, the asymptotic estimation variance of $g(\hat{\theta}_Y)$ is no greater than $g(\hat{\theta}_Z)$.
\label{repcorollary:1}
\end{repeat-corollary}

\noindent \textbf{Proof}: Under correct specification and standard regularity at $\theta^*$, the MLE is asymptotically normal with covariance equal to the Fisher Information as MLE Central Limit Theorem converges in distribution:

\begin{align*}
        \sqrt{n} (\hat{\theta}_Y - \theta^*) \xrightarrow{d} \mathcal{N}(0,{\mathcal{I}_Y(\theta^*)}^{-1}) \\
        \sqrt{n} (\hat{\theta}_Z - \theta^*) \xrightarrow{d} \mathcal{N}(0,{\mathcal{I}_Z(\theta^*)}^{-1}) 
\end{align*}

\noindent From \cref{theorem:1}, we have ${\mathcal{I}_Y(\theta)} \succeq {\mathcal{I}_Z(\theta)}$, if $A \succeq B \succ 0$ then $A^{-1} \preceq B^{-1}$ Loewner order reverses under inversion. So,

\begin{align*}
        {\mathcal{I}_Y(\theta^*)}^{-1} \preceq {\mathcal{I}_Z(\theta^*)}^{-1}
\end{align*}

\noindent The result so far is about quality of the parameter (weight) estimates. We are interested in the quality of our predictions. To make that final connection, we use the delta method. Consider any smooth function $g(\theta)$ - for instance, the target-class posterior $\tau (\theta) = \eta_c(x_0;\theta)$, where $x_0$ denotes an arbitrary fixed input. By the delta method, 

\begin{align*}
        \sqrt{n} (g(\hat{\theta}) - g(\theta_0)) \xrightarrow{d} \mathcal{N}(0,G\Sigma G^T) 
\end{align*}

\noindent where $G = \nabla_\theta g(\theta_0)$, $\theta_0$ is a parameter with non-singular Fisher information, and $\Sigma$ is the asymptotic covariance of the corresponding MLE. Therefore, each of the models can be written as:

\begin{align*}
        \sqrt{n} (\tau(\hat{\theta}_Y) - \tau(\theta^*)) \xrightarrow{d} \mathcal{N}(0,\nabla_{\tau}^T {\mathcal{I}_Y(\theta^*)}^{-1} \nabla_{\tau}) \\
        \sqrt{n} (\tau(\hat{\theta}_Z) - \tau(\theta^*)) \xrightarrow{d} \mathcal{N}(0,\nabla_{\tau}^T {\mathcal{I}_Z(\theta^*)}^{-1} \nabla_{\tau}) 
\end{align*}

\noindent From \cref{theorem:1}, we can show that,

\begin{align*}
        \nabla_{\tau}^T {\mathcal{I}_Y(\theta^*)}^{-1} \nabla_{\tau} \leq \nabla_{\tau}^T {\mathcal{I}_Z(\theta^*)}^{-1} \nabla_{\tau}
\end{align*}

\noindent So,

\vspace{-1em}
\begin{multline*}
\mathrm{Var}\!\left(\sqrt{n}\,(\tau(\hat{\theta}_Y) - \tau(\theta^*))\right) \\
\le \mathrm{Var}\!\left(\sqrt{n}\,(\tau(\hat{\theta}_Z) - \tau(\theta^*))\right)
\end{multline*}

Hence the estimates for $p_{\theta} (Y=c \mid x_0)$ are asymptotically tighter under multiclass
training. 

\qed

\subsection{Proof of Proposition 1}
\label{sec:proofs-p1}

\begin{repeat-proposition}[Softmax Expected Fisher Information Decomposition.]
Consider a softmax model with logits $f(x; \theta) \in \mathbb{R}^K$ and class probabilities $\eta = \text{softmax(f)}$. Let $J_f(x;\theta) = \partial f(x;\theta)/\partial \theta \in \mathcal{R}^{K \times p}$ denote the Jacobian of the logits with respect to the model parameters and let $e_c\in\mathbb R^{K}$ denote the $c$-th standard basis (one-hot) vector. Then for any fixed input $X=x$:
\begin{equation*}
  \mathcal{I}_Y(\theta \mid X=x)
  =J_f^T(\text{Diag}(\eta) - \eta \eta^T)J_f
\end{equation*}
is the conditional expected Fisher information for the multiclass likelihood, and
\begin{equation*}
  \mathcal{I}_Z (\theta \mid X=x) = J_f^T \left(\frac{\eta_c}{(1-\eta_c)} (e_c - \eta) {(e_c - \eta)}^T \right) J_f 
\end{equation*}
is the corresponding quantity for the collapsed-binary label. Hence, 
\begin{multline*}
\mathcal{I}_Y(\theta \mid X=x) = \mathcal{I}_Z(\theta \mid X=x) + \\
(1-\eta_c) J_f^T (\text{Diag}(\pi) - \pi \pi^T)J_f \succeq  \mathcal{I}_Z(\theta \mid X=x) 
\end{multline*}
where $\pi_k = \eta_k / (1- \eta_c)$ for $k \neq c$ and $\pi_c =0$.
\label{repproposition:1}
\end{repeat-proposition}

\noindent \textbf{Proof}: Let $f(x; \theta) \in \mathbb{R}^K$ be the logits and $\eta = \text{softmax(f)}$. For one observation $(X=x, Y=y)$ the log-likelihood for the datapoint is 

\begin{align*}
        \ell (\theta; y; x) = \log \eta_y = f_y - \log \sum_{j=1}^{K} e^{f_j}
\end{align*}

\noindent To get the conditional expected fisher information let $J_f(x;\theta) = \nabla_{\theta} f(x;\theta)$ be the Jacobian of the logits with respect to $\theta$. Then by the chain rule,

\begin{align*}
        s_Y (\theta; y; x) = \nabla_{\theta}\ell = {\left( \frac{\partial f}{\partial \theta}\right)}^T \nabla_f \ell = J_f^T \nabla_f \log \eta_y
\end{align*}

\noindent The softmax derivatives are, 

\begin{multline*}
         {\frac{\partial \eta_y}{\partial f_j}} = \eta_y (\delta_{yj} - \eta_j) \\ \implies \nabla_f \log \eta_y = \frac{1}{\eta_y} \nabla_f \eta_y = \delta_y - \eta = e_y - \eta
\end{multline*}

\noindent $\delta_{yj}$ is the Kronecker delta and $e_y$ is the $y$-th standard basis vector. So the score function can be written as 

\begin{align*}
         s_Y (\theta; y; x) = J_f^T (e_y - \eta)
\end{align*}

\noindent So the conditional expectation at a point $(X = x, Y = y)$ is 

\begin{align*}
         \mathcal{I}_Y (\theta \mid x) = \sum_{y=1}^k \eta_y [s_Y(\theta;y;x) {s_Y(\theta;y;x)}^T] \\ 
         \mathcal{I}_Y (\theta \mid x) = \sum_{y=1}^k \eta_y J_f^T (e_y - \eta) {(e_y - \eta)}^T J_f
\end{align*}

\noindent The inner sum is the covariance of the one-hot vector $e_Y$ with mean $\eta$

\begin{align*}
         \sum_{y=1}^k \eta_y e_y e_y^T -\eta \eta^T = \text{Diag}(\eta) - \eta \eta^T
\end{align*}

\noindent Therefore, 

\begin{align*}
         \mathcal{I}_Y (\theta \mid x) = J_f^T(\text{Diag}(\eta) - \eta \eta^T)J_f
\end{align*}

\noindent For the conditional expected fisher information for the collapsed Bernoulli case we start with the log-likelihood.

\begin{align*}
         \ell_Z (\theta; z; x) = z\log q + (1-z) \log (1-q)
\end{align*}

\noindent Where $q = q(x;\theta)$. Differentiating w.r.t. $\theta$ we get 

\begin{align*}
         \nabla_\theta \ell_Z = \left(\frac{z}{q} - \frac{1-z}{1-q}\right) \nabla_\theta q = \frac{z-q}{q(1-q)} \nabla_\theta q
\end{align*}

\noindent Therefore the score function for the collapsed Bernoulli is 

\begin{align*}
        s_Z (\theta; z; x) = \frac{z-q}{q(1-q)} \nabla_\theta q
\end{align*}

\noindent Therefore the score function for the collapsed Bernoulli is 

\begin{align*}
        s_Z (\theta; z; x) = \frac{z-q}{q(1-q)} \nabla_\theta q
\end{align*}

\noindent So the expected Fisher Information at $x$ is 

\begin{align*}
    \mathcal{I}_Z (\theta \mid X=x) = \mathbb{E}_\theta[s_Z(\theta) s_Z(\theta)^T \mid X=x] 
\end{align*}

\begin{align*}
    \mathcal{I}_Z (\theta \mid X=x) = \frac{\mathbb{E}_{\theta}[{(Z-q)}^2\mid x]}{q^2(1-q)^2} (\nabla_\theta q) {(\nabla_\theta q)}^T
\end{align*}

\noindent The numerator term is $\mathrm{Var}(Z \mid X =x) = \mathbb{E}[Z \mid X] - {(\mathbb{E}[Z \mid X])}^2 = q-q^2 = q(1-q)$ 

\begin{align*}
    \mathcal{I}_Z (\theta \mid X=x) = \frac{1}{q(1-q)} (\nabla_\theta q) {(\nabla_\theta q)}^T
\end{align*}

\noindent To get $\nabla_\theta q$, we take $q = \eta_c (f)$

\begin{align*}
         {\frac{\partial \eta_c}{\partial f_j}} = \eta_c (\delta_{cj} - \eta_j) \implies  \nabla_f \eta_c = \eta_c (e_c - \eta)
\end{align*}

\noindent Therefore, 

\begin{align*}
         \nabla_\theta q = {\left( \frac{\partial f}{\partial \theta}\right)}^T \nabla_f \eta_c = J_f^T (\eta_c (e_c - \eta))
\end{align*}

\noindent Since $q = \eta_c$

\begin{align*}
    \mathcal{I}_Z (\theta \mid X=x) = J_f^T \frac{1}{\eta_c(1-\eta_c)} (\eta_c (e_c - \eta))  (\eta_c {(e_c - \eta)}^T) J_f 
\end{align*}

\begin{align*}
    \mathcal{I}_Z (\theta \mid X=x) = J_f^T \left(\frac{\eta_c}{(1-\eta_c)} (e_c - \eta) {(e_c - \eta)}^T \right) J_f 
\end{align*}

\noindent To relate the two, express the class probabilities as
$\eta = [\,\eta_c,\,(1-\eta_c)\pi\,]$,
where $\pi_k = \eta_k/(1-\eta_c)$ for $k\neq c$ and
$\pi_c=0$.  Substituting this form into the multiclass curvature
in logit space gives

\begin{multline*}
\operatorname{Diag}(\eta)-\eta\eta^\top \\
  = \frac{\eta_c}{1-\eta_c}(e_c-\eta)(e_c-\eta)^\top
     +(1-\eta_c)\big(\operatorname{Diag}(\pi)-\pi\pi^\top\big).
\end{multline*}

Multiplying by $J_f^\top$ and $J_f$ on both sides yields the desired
decomposition:
\begin{multline*}
\mathcal{I}_Y(\theta\mid X=x) 
  = \mathcal{I}_Z(\theta\mid X=x) \\
    +(1-\eta_c)\,
      J_f^\top
      \big(\operatorname{Diag}(\pi)-\pi\pi^\top\big)
      J_f
  \succeq \mathcal{I}_Z(\theta\mid X=x).
\end{multline*}

The second term is positive semi--definite and quantifes the information lost by
collapsing all non--target classes into a single background label.

\qed

\subsection{Limitations for Classification Tasks}

While the proposed theory extends naturally to segmentation, it becomes less reliable in standard classification settings due to severe class imbalance. When $K$ classes are mapped into a single ``target vs.~rest'' binary task, the positive class often occupies a vanishing fraction of the data, yielding posterior probabilities $\eta_c(x;\theta)$ that are extremely close to $0$ or $1$. In this regime, the Bernoulli Fisher term $\eta_c(1-\eta_c)$ degenerates, leading to ill-conditioned or singular information matrices. As a result, the asymptotic efficiency and variance comparisons derived under balanced or well-behaved posteriors no longer hold. In practice, this means that for highly imbalanced classification datasets, coarsening amplifies numerical instability and the theoretical ordering $\mathcal{I}_Y(\theta)\succeq\mathcal{I}_Z(\theta)$ may not provide a useful description of estimator behavior.

\subsection{Note on Natural Images vs 3D Medical Images}

The observed strength of the proposed framework in 3D medical imaging tasks can be attributed to the high variability and structured heterogeneity of the \emph{background} in volumetric scans. Unlike 2D natural-image segmentation, where the ``non-target'' regions are often homogeneous and semantically uninformative (e.g., sky, wall, road), medical volumes contain multiple anatomical structures---organs, vessels, and tissues---each contributing distinct non-target gradients. This diversity amplifies the within-rest Fisher term in our decomposition, yielding a substantial gap between the multiclass and collapsed-binary information. Consequently, modeling each anatomical class preserves rich curvature directions in parameter space, improving statistical efficiency and convergence. In contrast, for dichotomous segmentation tasks in natural images or datasets with relatively uniform backgrounds, the variation is minimal, making the binary and multiclass formulations nearly equivalent.

\section{Heuristics for selecting auxiliary classes}
\label{sec:heuristics}

In practice, the choice of auxiliary support structures plays an important role in maximizing the benefits of {\methodology}. We outline several practical heuristics that can guide this selection.

\noindent \textbf{1. Select the organ containing the lesion.}  
The most effective auxiliary structure is typically the organ in which the lesion resides, as it fully encloses the target region. Including this structure allows the model to learn fine-grained variations in local tissue appearance, which directly improves boundary precision and reduces ambiguity.

\noindent \textbf{2. Include adjacent or surrounding organs.}  
Structures that spatially border the lesion serve as anatomical anchors. These surrounding organs provide contextual cues that help the model disambiguate lesion boundaries and reduce drift into neighboring regions. This is particularly useful for lesions located near organ interfaces.

\noindent \textbf{3. Add organs prone to false positives.}  
A third useful strategy is to incorporate organs whose appearance may cause confusion with the target lesion. These structures often contain components or textures that resemble the lesion class, leading to false positives in standard training. By explicitly modeling these tissues, the network learns discriminative cues that reduce such errors and improve robustness.

These heuristics offer a practical starting point for selecting auxiliary structures and can be adapted based on dataset characteristics, anatomical complexity, and known failure modes of baseline models.

\subsection{Limitations}

A key assumption of {\methodology} is the availability of auxiliary background structures. While such annotations are increasingly accessible in common modalities such as CT, MRI, and X-ray, often via large pretrained models (e.g., TotalSegmentator~\cite{d2024totalsegmentator}), this may not hold for less common modalities or domains lacking broad segmentation models. In these cases, obtaining auxiliary structures can be more challenging. To mitigate this, we suggest leveraging interactive segmentation methods to efficiently generate approximate background annotations, which, as demonstrated in our experiments, can still provide meaningful performance gains despite being noisy.

\subsection{Future Work on Heuristics}

While {\methodology} demonstrates consistent improvements across diverse settings, the choice of auxiliary structures remains an open question that warrants deeper theoretical and empirical investigation. In particular, it is unclear how the proximity and relevance of auxiliary structures influence performance. For example, when no structure directly encloses the target, it remains to be understood whether supervising nearby or partially overlapping structures is sufficient. Additionally, factors such as target size and anatomical context may modulate the observed gains, e.g., whether smaller lesions benefit more from organ-level supervision than larger ones. Addressing these questions would provide a more principled understanding of how to optimally select auxiliary structures and further improve the effectiveness of the proposed paradigm.

\section{Model Weights and Code}
\label{sec:code}
Code and pretrained weights can be found at this link \href{https://github.com/rachitsaluja/BackSplit}{github.com/rachitsaluja/BackSplit}. All models are implemented in PyTorch~\cite{paszke2019pytorch}, and inference can be performed directly using the nnU-Net~\cite{isensee2021nnu} library.

\section{Training Configuration and Hyperparameters}
\label{sec:params}

The following section summarizes the training configurations used for all models.

\subsection{U-Nets}

All U-Net~\cite{ronneberger2015u} models were trained using the nnU-Net~\cite{isensee2021nnu} 3D full-resolution configuration. A batch size of 2 and varying patch sizes were used across experiments, depending on dataset-specific memory requirements. All data were normalized according to imaging modality (CT or MRI) and subsequently resampled to the standardized nnU-Net resolution. The network backbone followed nnU-Net’s generic 3D U-Net design, employing InstanceNorm3d and LeakyReLU nonlinearities.

Training used nnU-Net’s default Dice + Cross-Entropy loss with deep supervision enabled. Optimization was performed with stochastic gradient descent (momentum 0.99), an initial learning rate of 0.01, weight decay of $3 \times 10^{-5}$, and the PolyLR learning rate schedule. Each model was trained for 1000 epochs, with 250 iterations per epoch. All experiments were conducted on an NVIDIA A100 GPU.

\subsection{ResEncU-Net}

The ResEncU-Net~\cite{isensee2024nnu} models follow the same training configuration as the U-Nets, differing only in network architecture. We use the Residual Encoder U-Net with the Medium configuration, as implemented in the nnU-Net framework. All optimization settings, normalization procedures, and training schedules remain identical to those described in the U-Net subsection.

\subsection{SegResNet}

For the SegResNet experiments, we use the MONAI~\cite{cardoso2022monai} SegResNet~\cite{myronenko20183d} implementation wrapped inside the nnU-Net framework, keeping all data preprocessing, normalization, and resampling steps identical to the nnU-Net. The primary differences lie in the network architecture and optimizer settings. Unlike the standard nnU-Net configuration, which uses SGD, SegResNet is optimized using Adam with a learning rate of \(1 \times 10^{-4}\), a weight decay of \(1 \times 10^{-5}\), an \(\varepsilon\) value of \(1 \times 10^{-5}\) and a PolyLR learning rate schedule. Models were trained using the default Dice + Cross-Entropy loss without deep supervision enabled. 

\begin{table}[t!]
\centering
\tablestyle{3pt}{1.3}
\begin{tabular}{r|cc}
\textbf{Auxiliary Task} & \textbf{Dice} $\uparrow$ & \textbf{Dice} $\uparrow$ \\
(Support Structure) & \textcircled{+}
7 clicks & \textcircled{+}10 clicks \\
\shline
\shline
\multicolumn{3}{c}{(a) KiTS23} \\
\hline
Kidney & 0.7420 & 0.7419 \\
Tumor & 0.5457 & 0.5428 \\
\shline
\shline
\multicolumn{3}{c}{(b) PANTHER-MR} \\
\hline
Pancreas & 0.6779 & 0.6823 \\
\shline
\shline
\multicolumn{3}{c}{(c) NSCLC-Rad} \\
\hline
Esophagus & 0.7155 & 0.7256 \\
Heart & 0.8408 & 0.8390 \\
Lung Left & 0.9606 & 0.9687 \\
Lung Right & 0.9627  & 0.9688 \\
Spinal Cord & 0.8244  & 0.8263 \\
\shline
\shline
\multicolumn{3}{c}{(d–e) AutoPET (MSWAL)} \\
\hline
Aorta & 0.8421 (0.9335) & 0.8506 (0.9335) \\
Gall Bladder & 0.6765 (0.8003) & 0.6750 (0.8003) \\
Kidney Left & 0.9333 (0.9585) & 0.9334 (0.9585) \\
Kidney Right & 0.9416 (0.9589) & 0.9396 (0.9589) \\
Liver & 0.9388 (0.9662) & 0.9360 (0.9662) \\
Pancreas & 0.7228 (0.8497) & 0.7374 (0.8497) \\
Postcava & 0.7648 (0.7724) & 0.7804 (0.7724) \\
Spleen & 0.9379 (0.9655) & 0.9353 (0.9655) \\
Stomach & 0.8718 (0.8895) & 0.8791 (0.8895) \\
\shline
\shline
\end{tabular}   
\caption{Simulated interactive segmentation results using 7 and 10 user clicks (\textcircled{+}) across five datasets for generating support structures. For KiTS, PANTHER-MR, and NSCLC-Rad, ground-truth masks were used to randomly sample 7 or 10 seed points, followed by segmentation using nnInteractive. For AutoPET and MSWAL, the seed points were sampled from model-derived organ segmentations.}
\label{tab:nni_results_aux}
\vspace{-1.5em}
\end{table}

\subsection{SwinUNETR}

For the transformer-based experiments (as shown in \cref{sec:transformers}), we employ the 3D SwinUNETR~\cite{hatamizadeh2021swin} architecture from MONAI~\cite{cardoso2022monai}, integrated into the nnU-Net framework while keeping all preprocessing, normalization, and resampling procedures consistent with the standard nnU-Net pipeline. The main differences arise in the network architecture and optimization strategy.

SwinUNETR is trained using the AdamW optimizer with an initial learning rate of \(8 \times 10^{-4}\), a weight decay of \(0.01\), and an \(\varepsilon\) value of \(1 \times 10^{-5}\). To better suit transformer training dynamics, we replace nnU-Net’s default PolyLR schedule with a CosineAnnealingLR scheduler, setting \(T_{\max}\) to the total number of training epochs and a minimum learning rate of \(1 \times 10^{-6}\).

All other training settings—including loss function (Dice + Cross-Entropy), batch size, and epoch count—are kept consistent with the U-Net and SegResNet configurations unless otherwise noted.

\subsection{Minimal Parameter Overhead of {\methodology}}

An advantage of {\methodology} is its minimal parameter overhead. Even in the most demanding setting—where the largest number of auxiliary classes is added—the increase in model parameters is approximately \(0.02\%\). This negligible overhead ensures that {\methodology} remains practical, adding virtually no computational burden while providing strong performance gains.

\begin{table}[t!]
\centering
\tablestyle{3pt}{1.3}
\begin{tabular}{r|cc}
\textbf{Auxiliary Task} & \textbf{Dice} $\uparrow$ & \textbf{Dice} $\uparrow$ \\
(Support Structure) & TotalSegmentator & VIBESegmentator \\
\shline
\shline
\multicolumn{3}{c}{(a–b) AutoPET (MSWAL)} \\
\hline
Aorta & 0.9228 (0.9216) & 0.7748 (0.8211) \\
Gall Bladder & 0.8307 (0.7612) & 0.3875 (0.4609) \\
Kidney Left & 0.9192 (0.9427) & 0.9266 (0.8799) \\
Kidney Right & 0.9214 (0.9433) & 0.9282 (0.8723) \\
Liver & 0.9761 (0.9719) & 0.571 (0.8673) \\
Pancreas & 0.8834 (0.8841) & 0.6592 (0.697) \\
Postcava & 0.9062 (0.8614) & 0.8275 (0.7092) \\
Spleen & 0.9631 (0.9578) & 0.848 (0.8284) \\
Stomach & 0.9473 (0.9377) & 0.8728 (0.8036) \\
\shline
\shline
\end{tabular}   
\caption{Auxiliary segmentation performance from large pretrained models on AutoPET and MSWAL. Reported Dice scores reflect organ segmentations generated by TotalSegmentator and VIBESegmentator, with values in parentheses indicating corresponding performance on MSWAL. These automatically derived masks are used as support structures for {\methodology}.}
\label{tab:zoo_aux}
\vspace{-1.5em}
\end{table}

\section{Interactive Segmentation Performance}
\label{sec:nni_perf}

Table~\ref{tab:nni_results_aux} reports the segmentation quality of auxiliary support structures generated using nnInteractive~\cite{isensee2025nninteractive} with 7 and 10 positive clicks. Although the resulting masks exhibit only modest accuracy, they are sufficiently informative for {\methodology} to produce strong performance gains in downstream lesion segmentation tasks. This demonstrates that {\methodology} remains effective even when auxiliary labels are noisy and derived from lightweight interactive segmentation rather than precise manual annotations.

\section{Large Models Segmentation Performance}
\label{sec:lm_perf}

To further assess the robustness of {\methodology}, we evaluate auxiliary segmentations produced by large pretrained models—TotalSegmentator~\cite{d2024totalsegmentator} and VIBESegmentator~\cite{graf2025vibesegmentator}—on the AutoPET and MSWAL datasets. These models are widely used for comprehensive whole-body and abdominal organ parsing and provide a strong reference point for high-quality automatic segmentation. We compare their outputs with our model-derived segmentations obtained from a U-Net trained on the AbdomenAtlasMini1.0~\cite{qu2023abdomenatlas} dataset. As shown in Table~\ref{tab:zoo_aux}, the pretrained models achieve strong organ segmentation accuracy on AutoPET and MSWAL.

\begin{table}[t]
\centering
\tablestyle{2pt}{1.5}
\begin{tabular}{r|ccc|ccc}
& \multicolumn{3}{c|}{\textbf{Patch Size Variation 1}} 
& \multicolumn{3}{c}{\textbf{Patch Size Variation 2}} \\
\textbf{Method} 
& \textbf{Dice} $\uparrow$ & \textbf{HD-95} $\downarrow$ & \textbf{NSD} $\uparrow$
& \textbf{Dice} $\uparrow$ & \textbf{HD-95} $\downarrow$ & \textbf{NSD} $\uparrow$ \\
\shline
\shline
\multicolumn{7}{c}{(a) Dataset: KiTS23 with Target Class = \{Cyst\}} \\
\shline
Patch Size Value & \multicolumn{3}{c|}{[128,128,128]} &  \multicolumn{3}{c}{[160,128,128]} \\
\shline
Regular Training
& 0.2033 & 425.3273 & 0.1906 
& 0.2117 & 422.4271 &  0.1981 \\
\rlightgray
\methodology 
& 0.5297 & 249.5371 & 0.6703 
& 0.5295 & 251.8412 & 0.6732 \\
\shline
\shline
\multicolumn{7}{c}{(b) Dataset: PANTHER-MR with Target Class = \{Tumor\}} \\
\shline
Patch Size Value & \multicolumn{3}{c|}{[48,192,224]} &  \multicolumn{3}{c}{[48,224,224]} \\
\shline
Regular Training
& 0.3828 & 157.3470 & 0.2320
& 0.4409 & 152.3861  & 0.2623 \\
\rlightgray
\methodology 
& 0.4906 & 54.2010 & 0.2796 
& 0.4732 & 52.0249 & 0.2718 \\
\shline
\shline
\multicolumn{7}{c}{(c) Dataset: NSCLC-Radiomics with Target Class = \{GTV\}} \\
\shline
Patch Size Value & \multicolumn{3}{c|}{[64,192,192]} &  \multicolumn{3}{c}{[32,160,192]} \\
\shline
Regular Training
& 0.5279 & 140.2698 & 0.4049 
& 0.4984 & 164.0937 & 0.3771 \\
\rlightgray
\methodology 
&  0.5862 & 124.2557 & 0.4533
& 0.5724 &  123.7844 & 0.4431 \\
\shline
\shline
\end{tabular}
\caption{Single-fold evaluation of {\methodology} vs. regular training using a U-Net backbone across two patch-size configurations (batch size fixed at 2). Results are shown for (a) KiTS23, (b) PANTHER-MR, and (c) NSCLC-Radiomics. Across all datasets and both patch-size settings, {\methodology} consistently outperforms regular training on Dice, HD-95, and NSD metrics, demonstrating robustness to changes in patch size.}
\label{tab:patch_size}
\vspace{-1.5em}
\end{table}

\section{{\methodology} Performance during different training settings}
\label{sec:settings}

In this section, we evaluate the effectiveness of {\methodology} across different training conditions and architectural variations to further demonstrate its robustness as a general paradigm.

\subsection{Evaluating {\methodology} Under Varying Batch Sizes}

To assess the robustness of {\methodology}, we first evaluate its performance under varying batch sizes. Given typical GPU memory constraints for 3D medical segmentation models, we consider realistic batch sizes of 2 and 4, and conduct experiments on the KiTS23~\cite{heller2023kits21}, PANTHER~\cite{betancourt_tarifa_2025_15192302}, and NSCLC-Radiomics~\cite{aerts2015data} datasets. In all cases, {\methodology} consistently outperforms regular training, demonstrating its effectiveness irrespective of batch size (as shown in \cref{tab:batch_size}). All experiments were performed on a single fold using U-Net~\cite{ronneberger2015u} with the nnU-Net framework~\cite{isensee2021nnu}.

\begin{table}[t]
\centering
\tablestyle{2pt}{1.5}
\begin{tabular}{r|ccc|ccc}
& \multicolumn{3}{c|}{\textbf{Batch Size = 2}} 
& \multicolumn{3}{c}{\textbf{Batch Size = 4}} \\
\textbf{Method} 
& \textbf{Dice} $\uparrow$ & \textbf{HD-95} $\downarrow$ & \textbf{NSD} $\uparrow$
& \textbf{Dice} $\uparrow$ & \textbf{HD-95} $\downarrow$ & \textbf{NSD} $\uparrow$ \\
\shline
\shline
\multicolumn{7}{c}{(a) Dataset: KiTS23 with Target Class = \{Cyst\}} \\
\shline
Regular Training
& 0.2033 & 425.3273 & 0.1906 
& 0.2630 & 404.6592 & 0.2624 \\
\rlightgray
\methodology 
& 0.5297 & 249.5371 & 0.6703 
& 0.5261 & 249.6824 & 0.6716 \\
\shline
\shline
\multicolumn{7}{c}{(b) Dataset: PANTHER-MR with Target Class = \{Tumor\}} \\
\shline
Regular Training
& 0.3828 & 157.3470 & 0.2320 
& 0.3636 & 202.4875 & 0.2129 \\
\rlightgray
\methodology 
& 0.4906 & 54.2010 & 0.2796 
& 0.4553 & 59.1720 & 0.2696 \\
\shline
\shline
\multicolumn{7}{c}{(c) Dataset: NSCLC-Radiomics with Target Class = \{GTV\}} \\
\shline
Regular Training
& 0.5279 & 140.2698 & 0.4049 
& 0.5530 & 136.1963 & 0.4270  \\
\rlightgray
\methodology 
&  0.5862 & 124.2557 & 0.4533
& 0.5549 & 127.0088 & 0.4353 \\
\shline
\shline
\end{tabular}
\caption{Single-fold evaluation of {\methodology} vs. regular training using a U-Net backbone across two batch sizes (2 and 4). Results are shown for (a) KiTS23, (b) PANTHER-MR, and (c) NSCLC-Radiomics. In all datasets and for both batch-size settings, {\methodology} consistently outperforms regular training across Dice, HD-95, and NSD metrics, demonstrating robustness to changes in batch size.}
\label{tab:batch_size}
\vspace{-1.5em}
\end{table}

\subsection{Evaluating {\methodology} Under Varying Patch Sizes}

To further evaluate the robustness of {\methodology}, we examine its performance under varying patch sizes for each dataset. Patch size is a critical factor in 3D medical segmentation due to differences in anatomical scale and memory constraints. We conduct experiments on the KiTS23~\cite{heller2023kits21}, PANTHER~\cite{betancourt_tarifa_2025_15192302}, and NSCLC-Radiomics~\cite{aerts2015data} datasets using multiple patch-size configurations. Across all settings, {\methodology} consistently outperforms regular training, demonstrating its stability with respect to patch-size variation (as shown in \cref{tab:patch_size}). All experiments were performed on a single fold using U-Net~\cite{ronneberger2015u} within the nnU-Net framework~\cite{isensee2021nnu} keeping batch size constant, at a size of two.

\subsection{{\methodology} with and without Deep Supervision}

We also assess the impact of deep supervision~\cite{lee2015deeply}, a mechanism used in the nnU-Net framework to provide intermediate supervision from the layers and enhance performance in 3D segmentation networks. For each dataset, we compare models trained with and without deep supervision while keeping all other training settings fixed. In both configurations, {\methodology} consistently outperforms standard training, indicating that its benefits are complementary to deep supervision rather than dependent on it (as shown in \cref{tab:deep_super}). All experiments were conducted on a single fold using U-Net~\cite{ronneberger2015u} within the nnU-Net framework~\cite{isensee2021nnu}.

\begin{table}[t]
\centering
\tablestyle{2pt}{1.5}
\begin{tabular}{r|ccc|ccc}
& \multicolumn{3}{c|}{\textbf{with}} 
& \multicolumn{3}{c}{\textbf{without}} \\
& \multicolumn{3}{c|}{\textbf{Deep Supervision~\cite{lee2015deeply}}} 
& \multicolumn{3}{c}{\textbf{Deep Supervision~\cite{lee2015deeply}}} \\
\textbf{Method} 
& \textbf{Dice} $\uparrow$ & \textbf{HD-95} $\downarrow$ & \textbf{NSD} $\uparrow$
& \textbf{Dice} $\uparrow$ & \textbf{HD-95} $\downarrow$ & \textbf{NSD} $\uparrow$ \\
\shline
\shline
\multicolumn{7}{c}{(a) Dataset: KiTS23 with Target Class = \{Cyst\}} \\
\shline
Regular Training
& 0.2033 & 425.3273 & 0.1906 
& 0.1339 & 441.7240 & 0.1300 \\
\rlightgray
\methodology 
& 0.5297 & 249.5371 & 0.6703 
& 0.4758 & 281.8879 & 0.6137 \\
\shline
\shline
\multicolumn{7}{c}{(b) Dataset: PANTHER-MR with Target Class = \{Tumor\}} \\
\shline
Regular Training
& 0.3828 & 157.3470 & 0.2320 
&  0.4412 & 179.4845 & 0.2590 \\
\rlightgray
\methodology 
& 0.4906 & 54.2010 & 0.2796 
& 0.4776 & 46.5688 & 0.2905 \\
\shline
\shline
\multicolumn{7}{c}{(c) Dataset: NSCLC-Radiomics with Target Class = \{GTV\}} \\
\shline
Regular Training
& 0.5279 & 140.2698 & 0.4049 
& 0.5059 & 158.2856 & 0.3765  \\
\rlightgray
\methodology 
&  0.5862 & 124.2557 & 0.4533
& 0.5829  & 115.2792 & 0.4564 \\
\shline
\shline
\end{tabular}
\caption{Single-fold evaluation of {\methodology} with and without Deep Supervision~\cite{lee2015deeply} using a U-Net backbone. Results are shown for (a) KiTS23, (b) PANTHER-MR, and (c) NSCLC-Radiomics. Across all datasets and under both training conditions, {\methodology} consistently outperforms regular training on Dice, HD-95, and NSD metrics, demonstrating that its benefits hold regardless of whether Deep Supervision is employed.}
\label{tab:deep_super}
\vspace{-1em}
\end{table}

\begin{table}[t]
\centering
\tablestyle{2pt}{1.5}
\begin{tabular}{r|ccc|ccc}
& \multicolumn{3}{c|}{\textbf{3D U-Net~\cite{isensee2021nnu, ronneberger2015u}}} 
& \multicolumn{3}{c}{\textbf{2D U-Net~\cite{isensee2021nnu, ronneberger2015u}}} \\
\textbf{Method} 
& \textbf{Dice} $\uparrow$ & \textbf{HD-95} $\downarrow$ & \textbf{NSD} $\uparrow$
& \textbf{Dice} $\uparrow$ & \textbf{HD-95} $\downarrow$ & \textbf{NSD} $\uparrow$ \\
\shline
\shline
\multicolumn{7}{c}{(a) Dataset: KiTS23 with Target Class = \{Cyst\}} \\
\shline
Regular Training
& 0.2033 & 425.3273 & 0.1906 
& 0.3234 & 352.3277 & 0.4231 \\
\rlightgray
\methodology 
& 0.5297 & 249.5371 & 0.6703 
& 0.4111 & 313.9846 & 0.5126 \\
\shline
\shline
\multicolumn{7}{c}{(b) Dataset: PANTHER-MR with Target Class = \{Tumor\}} \\
\shline
Regular Training
& 0.3828 & 157.3470 & 0.2320 
& 0.2349 & 265.2399 & 0.1312 \\
\rlightgray
\methodology 
& 0.4906 & 54.2010 & 0.2796 
& 0.3611 & 137.2006 & 0.2031 \\
\shline
\shline
\multicolumn{7}{c}{(c) Dataset: NSCLC-Radiomics with Target Class = \{GTV\}} \\
\shline
Regular Training
& 0.5279 & 140.2698 & 0.4049 
& 0.4695 & 102.8473 & 0.3676 \\
\rlightgray
\methodology 
&  0.5862 & 124.2557 & 0.4533
& 0.4753 & 125.5158 & 0.3728 \\
\shline
\shline
\end{tabular}
\caption{Single-fold evaluation of {\methodology} using 3D and 2D U-Net architectures on (a) KiTS23, (b) PANTHER-MR, and (c) NSCLC-Radiomics. Across all datasets and for both 3D and 2D models, {\methodology} consistently outperforms regular training on Dice, HD-95, and NSD metrics. These results demonstrate that the benefits of the BackSplit paradigm hold irrespective of network dimensionality.}
\label{tab:2d}
\vspace{-1.5em}
\end{table}

\subsection{{\methodology} Performance on 2D U-Net Architectures}

We additionally evaluate {\methodology} using both 3D U-Net and 2D U-Net architectures to assess its effectiveness across different network dimensionalities. While 3D models typically capture richer volumetric context and 2D models offer computational efficiency, {\methodology} yields consistent performance improvements in both settings. This demonstrates that the paradigm is not tied to a specific architectural dimensionality and remains effective whether the model processes full 3D volumes or individual 2D slices (as shown in \cref{tab:2d}). All experiments were performed on a single fold using implementations with the nnU-Net framework~\cite{isensee2021nnu} using their “3d\_fullres” and “2d” configurations.

\subsection{{\methodology} Performance on Transformer-based Architectures}
\label{sec:transformers}

We also evaluate {\methodology} on transformer-based architectures, despite prior work indicating that such models often underperform compared to convolutional counterparts in medical image segmentation~\cite{isensee2024nnu}. In particular, we employ a 3D SwinUNETR~\cite{hatamizadeh2021swin}. Even in this setting, {\methodology} yields consistent improvements over standard training (as shown in \cref{tab:transformer_results}), demonstrating that its benefits extend beyond convolutional models and remain effective for transformer-based architectures. However, as noted in the literature it does not perform as well as U-Nets.

\begin{table}[t]
\centering
\tablestyle{2pt}{1.5}
\begin{tabular}{r|ccc|ccc}
& \multicolumn{3}{c|}{\textbf{3D U-Net~\cite{isensee2021nnu, ronneberger2015u}}} 
& \multicolumn{3}{c}{\textbf{3D SwinUNETR~\cite{hatamizadeh2021swin}}} \\
\textbf{Method} 
& \textbf{Dice} $\uparrow$ & \textbf{HD-95} $\downarrow$ & \textbf{NSD} $\uparrow$
& \textbf{Dice} $\uparrow$ & \textbf{HD-95} $\downarrow$ & \textbf{NSD} $\uparrow$ \\
\shline
\shline
\multicolumn{7}{c}{(a) Dataset: KiTS23 with Target Class = \{Cyst\}} \\
\shline
Regular Training
& 0.2033 & 425.3273 & 0.1906 
& 0.1336 & 438.8709 & 0.1035 \\
\rlightgray
\methodology 
& 0.5297 & 249.5371 & 0.6703 
& 0.3863 & 309.6816 & 0.5010 \\
\shline
\shline
\multicolumn{7}{c}{(b) Dataset: PANTHER-MR with Target Class = \{Tumor\}} \\
\shline
Regular Training
& 0.3828 & 157.3470 & 0.2320 
& 0.3113 & 163.1136 & 0.1480 \\
\rlightgray
\methodology 
& 0.4906 & 54.2010 & 0.2796 
& 0.3705 & 59.1148 & 0.1926 \\
\shline
\shline
\multicolumn{7}{c}{(c) Dataset: NSCLC-Radiomics with Target Class = \{GTV\}} \\
\shline
Regular Training
& 0.5279 & 140.2698 & 0.4049 
& 0.4936  & 154.8597 & 0.3674 \\
\rlightgray
\methodology 
&  0.5862 & 124.2557 & 0.4533
& 0.5344 & 131.8715 & 0.3942 \\
\shline
\shline
\end{tabular}
\caption{Single-fold evaluation of {\methodology} on 3D U-Net and transformer-based (3D SwinUNETR) architecture across (a) KiTS23, (b) PANTHER-MR, and (c) NSCLC-Radiomics. In all datasets and for both architectures, {\methodology} consistently outperforms regular training on Dice, HD-95, and NSD metrics. These results demonstrate that the {\methodology} paradigm generalizes beyond CNNs and remains effective for transformer-based segmentation models.}
\label{tab:transformer_results}
\vspace{-1.5em}
\end{table}

\section{Empirical Analysis of {\methodology} with Increasing Auxiliary Classes}
\label{sec:K_class}

In this section, we analyze how performance metrics change when the number of auxiliary classes used in {\methodology} training is increased linearly, and compare these results against standard training.

For this experiment, we use the AutoPET~\cite{gatidis2022whole} dataset and incrementally add auxiliary classes from the AbdomenAtlas1.0Mini~\cite{qu2023abdomenatlas} dataset (with the left and right kidneys added simultaneously as two separate classes). This yields seven distinct models, in addition to the regular training model and the {\methodology} model containing all auxiliary structures, as shown in the main paper. All models were trained for a single fold using the U-Net~\cite{ronneberger2015u} architecture within the nnU-Net framework~\cite{isensee2021nnu}.

\begin{figure}[t]
  \centering
      \includegraphics[width=0.99\linewidth]{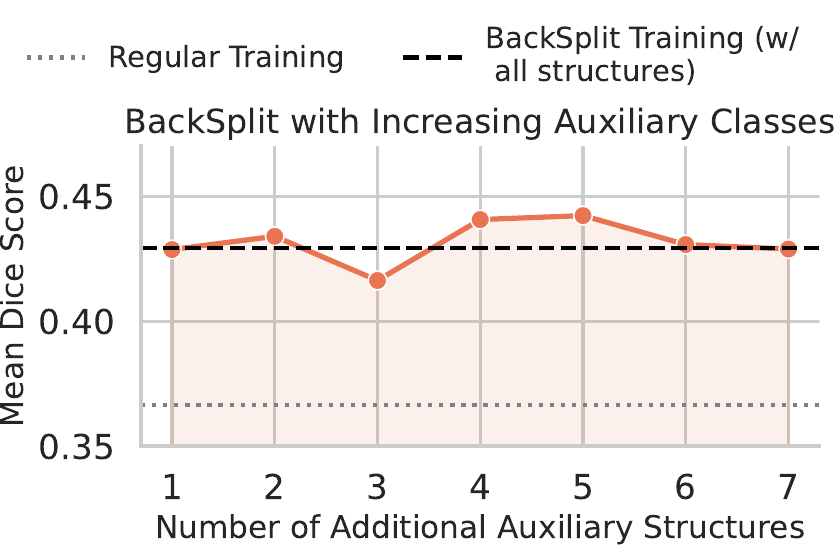}
   \caption{Adding even a single auxiliary structure yields an immediate improvement over regular training, and although incorporating multiple structures introduces mixed behavior with both gains and small drops, performance remains consistently well above the regular-training baseline. This experiment is conducted on the AutoPET dataset by incrementally adding auxiliary classes from AbdomenAtlas1.0Mini using a U-Net backbone.}
   \label{fig:vsNumClasses}
   % \vspace{-1em}
\end{figure}

We observe that the effect of {\methodology} is almost immediate: even adding a single auxiliary structure yields a notable performance gain (as shown in \cref{fig:vsNumClasses}). This is intuitive, as the inclusion of even one anatomical structure provides a meaningful contextual anchor for identifying lesions. Although adding multiple structures shows a mix of positive and negative trends whose underlying causes remain unclear, {\methodology} consistently outperforms regular training across all settings. As noted in \cref{sec:Conclusion}, identifying which support structures are most beneficial remains an important direction for future work.

\section{{\methodology} vs. Virtual Classes}
\label{sec:virtual_class}

We also conduct a brief comparison against the virtual classes paradigm~\cite{chen2018virtual}, which has been explored in classification tasks by adding an additional class only at the softmax layer during training. This approach is hypothesized to encourage more discriminative feature embeddings and thereby improve performance. To emulate this idea in our setting, we train a U-Net with an added empty class, effectively mimicking the virtual class formulation.

We observe a similar effect in which the virtual class formulation yields a modest increase in performance metrics; however, these gains remain substantially lower than those achieved with {\methodology}, as shown in \cref{tab:virtual_class}. These results suggest that the virtual class approach primarily encourages more discriminative features compared to standard training, whereas {\methodology} provides improved predictions through reduced variance and richer anatomical context.

From a practical implementation standpoint, we again use U-Net~\cite{ronneberger2015u} within the nnU-Net framework~\cite{isensee2021nnu} to demonstrate this effect on the KiTS23~\cite{heller2023kits21} dataset for a single fold. The virtual class is implemented by adding an additional prediction channel without providing corresponding labels in the dataset, and performance is evaluated solely on the cyst label.

\begin{table}[t]
\centering
\tablestyle{8pt}{1.2}
\begin{tabular}{r|ccc}
& \multicolumn{3}{c}{\textbf{KiTS23}} \\
\textbf{Method} 
& \textbf{Dice} $\uparrow$ & \textbf{HD-95} $\downarrow$ & \textbf{NSD} $\uparrow$ \\
\shline
\shline
Regular Training
&0.2033 & 425.3273 & 0.1906 \\
Training with Virtual Class
& 0.2469 & 414.6015 & 0.2467  \\
\rlightgray
\methodology 
& 0.5297 & 249.5371 & 0.6703  \\
\shline
\shline
\end{tabular}
\caption{Single-fold evaluation on KiTS23 (Target = Cyst) using a U-Net backbone. {\methodology} substantially outperforms both regular training and the Virtual Class paradigm across all metrics.}
\label{tab:virtual_class}
% \vspace{-1.5em}
\end{table}

\begin{figure}[t]
  \centering
  \vspace{-1em}
      \includegraphics[width=0.99\linewidth]{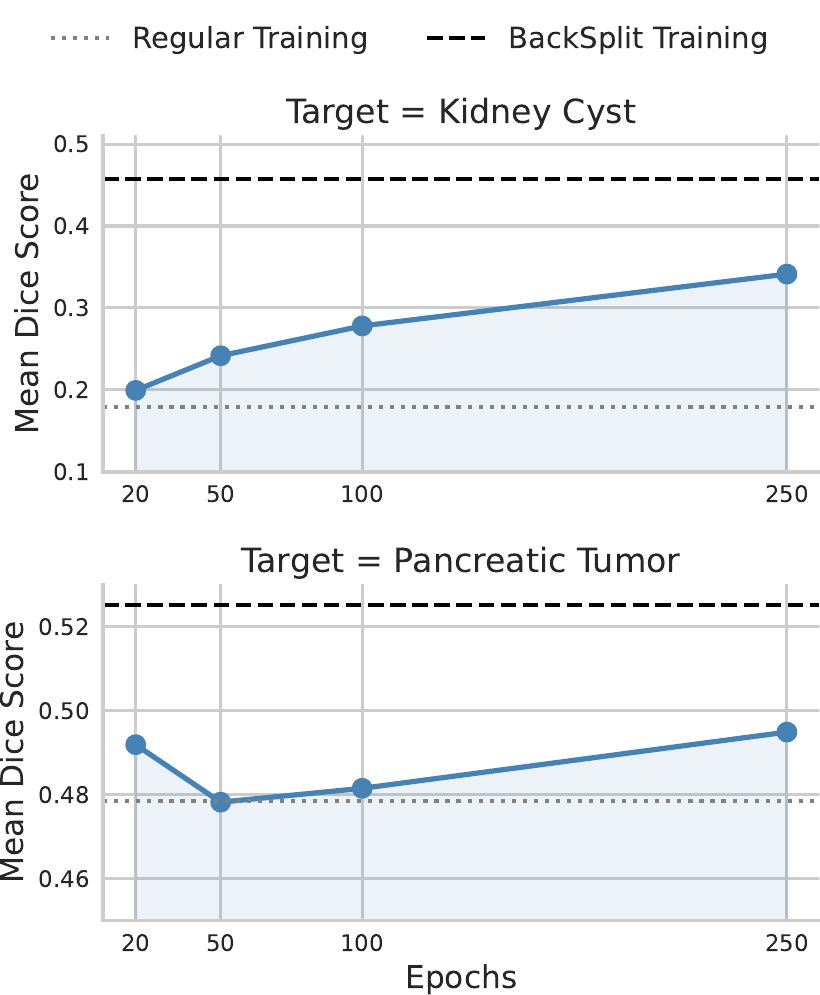}
   \caption{\textbf{(Top)} Effect of fine-tuning with {\methodology} on a pretrained binary model for kidney cyst segmentation (KiTS23). Mean Dice steadily improves with training epochs, approaching full {\methodology} performance. \textbf{(Bottom)} Similar trend observed for pancreatic tumor segmentation (PANTHER-MR), where fine-tuning progressively narrows the gap between regular and full {\methodology}.}
   \label{fig:ft_ext1}
   \vspace{-1.5em}
\end{figure}

\begin{figure}[t]
  \centering
  \vspace{-1em}
      \includegraphics[width=0.99\linewidth]{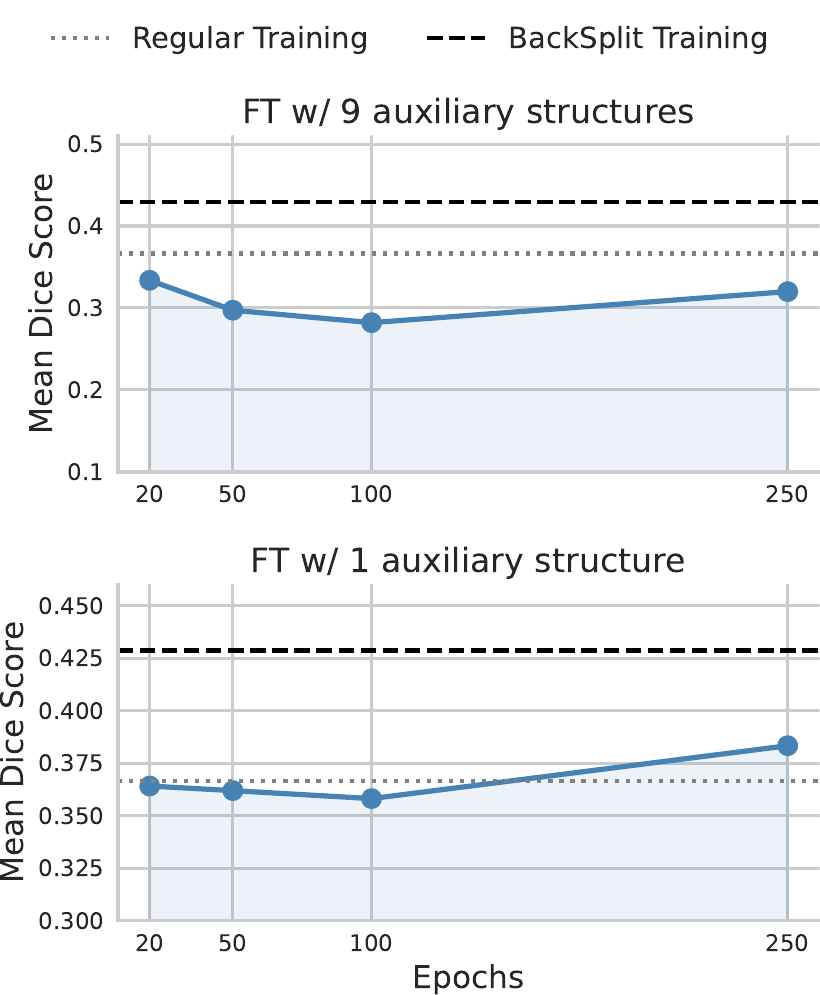}
   \caption{\textbf{(Top)} Fine-tuning on AutoPET with 9 auxiliary support structures. When many auxiliary labels are present, the model tends to learn the “easier’’ auxiliary classes first, delaying progress on the harder lesion target. Under a limited training budget (250 epochs), the model does not sufficiently reach the target class and fails to surpass the regular-training baseline. \textbf{(Bottom)} Fine-tuning with only 1 auxiliary structure avoids this and mirrors the behavior seen in KiTS23 and PANTHER, where performance steadily improves with training epochs.}
   \label{fig:ft_ext2}
   \vspace{-1.5em}
\end{figure}

\section{Extended Analysis of {\methodology} under Fine-Tuning}
\label{sec:ext_FT}

We further analyze the behavior of {\methodology} in the fine-tuning setting, providing additional guidance for readers on how to apply the paradigm effectively for improved performance.

{\methodology} is most effective in settings with a limited number of support structures. We demonstrate this empirically through three experiments on the KiTS23~\cite{heller2023kits21}, PANTHER~\cite{betancourt_tarifa_2025_15192302}, and AutoPET~\cite{gatidis2022whole} datasets (the latter under two configurations). As shown in the main paper, KiTS23 exhibits clear improvements (\cref{fig:ft_datamix} Left and \cref{fig:ft_ext1} Top), and PANTHER displays a similar trend (as shown in \cref{fig:ft_ext1} Bottom). However, when fine-tuning AutoPET with {\methodology}, the model struggles to surpass the baseline performance (as shown in \cref{fig:ft_ext2} Top).

We attribute this weaker performance to the experimental setup used for AutoPET, where {\methodology} incorporated a large number of support structures (nine in total). In such cases, the model tends to learn “easier” auxiliary labels first, delaying progress on the harder target label. With a training budget of only 250 epochs, the model does not sufficiently progress to learning the target class, which leads to the observed underperformance.

To mitigate this effect, we fine-tune the model using only a subset of support structures rather than all nine throughout training. Under this setting, AutoPET follows the same trend observed for KiTS23 and PANTHER (as shown in \cref{fig:ft_ext2} Bottom). This provides a practical way to manage training dynamics when dealing with a large number of auxiliary structures. Here, we just add one additional support structure.

All experiments in this section are conducted using U-Net~\cite{ronneberger2015u} via the nnU-Net~\cite{isensee2021nnu} framework. For \cref{fig:ft_ext1}, we report full 5-fold cross-validation results. In contrast, \cref{fig:ft_ext2} presents results from a single fold, as AutoPET is a substantially larger dataset and requires significantly more computational resources. This experiment is included primarily to provide empirical support for the claims made in this section. Additionally, following standard practice, we fine-tune the models using a lower initial learning rate of \(1 \times 10^{-3}\), compared to the typical \(1 \times 10^{-2}\) used in nnU-Net training.

As initialization from pretrained models becomes increasingly common in medical image segmentation (e.g., TotalSegmentator~\cite{d2024totalsegmentator}, MultiTalent~\cite{ulrich2023multitalent}), we additionally evaluate {\methodology} in this setting. Specifically, we initialize from the TotalSegmentator-CT model and fine-tune using the training protocol described above on the KiTS23 and NSCLC-Radiomics datasets. Across both datasets, {\methodology} continues to yield consistent performance improvements over standard fine-tuning, demonstrating that its benefits extend beyond training from scratch and remain effective when applied on top of strong pretrained representations(as shown in \cref{tab:ft-ts}).

\begin{table}[t]
\centering
\tablestyle{6pt}{1.2}
\begin{tabular}{r|cc|cc}
& \multicolumn{2}{c}{\textbf{Random Init}} & \multicolumn{2}{c}{\textbf{TS-CT~\cite{d2024totalsegmentator} Init}} \\
\textbf{Method} 
& \textbf{Regular}  & \textbf{{\methodology}}  &  \textbf{Regular}  & \textbf{{\methodology}} \\
\shline
\shline
KiTS23
&0.2033 & 0.5297 & 0.36 & 0.4072 \\
NSCLC-Rad
& 0.5279 & 0.5862 & 0.4716 & 0.5004  \\
\shline
\shline
\end{tabular}
\caption{Comparison of Dice scores for regular training and {\methodology} under different initialization strategies: random initialization and pretrained TotalSegmentator-CT (TS-CT) initialization. Results are reported for KiTS23 and NSCLC-Radiomics. {\methodology} consistently improves performance over regular training in both settings, demonstrating that its benefits persist when fine-tuning from strong pretrained models as well as when training from scratch.}
\label{tab:ft-ts}
\vspace{-1.5em}
\end{table}

\section{Qualitative Performance}
\label{sec:qualitative_perf}

\begin{figure*}[t!]
  \centering
   \includegraphics[width=0.8\linewidth]{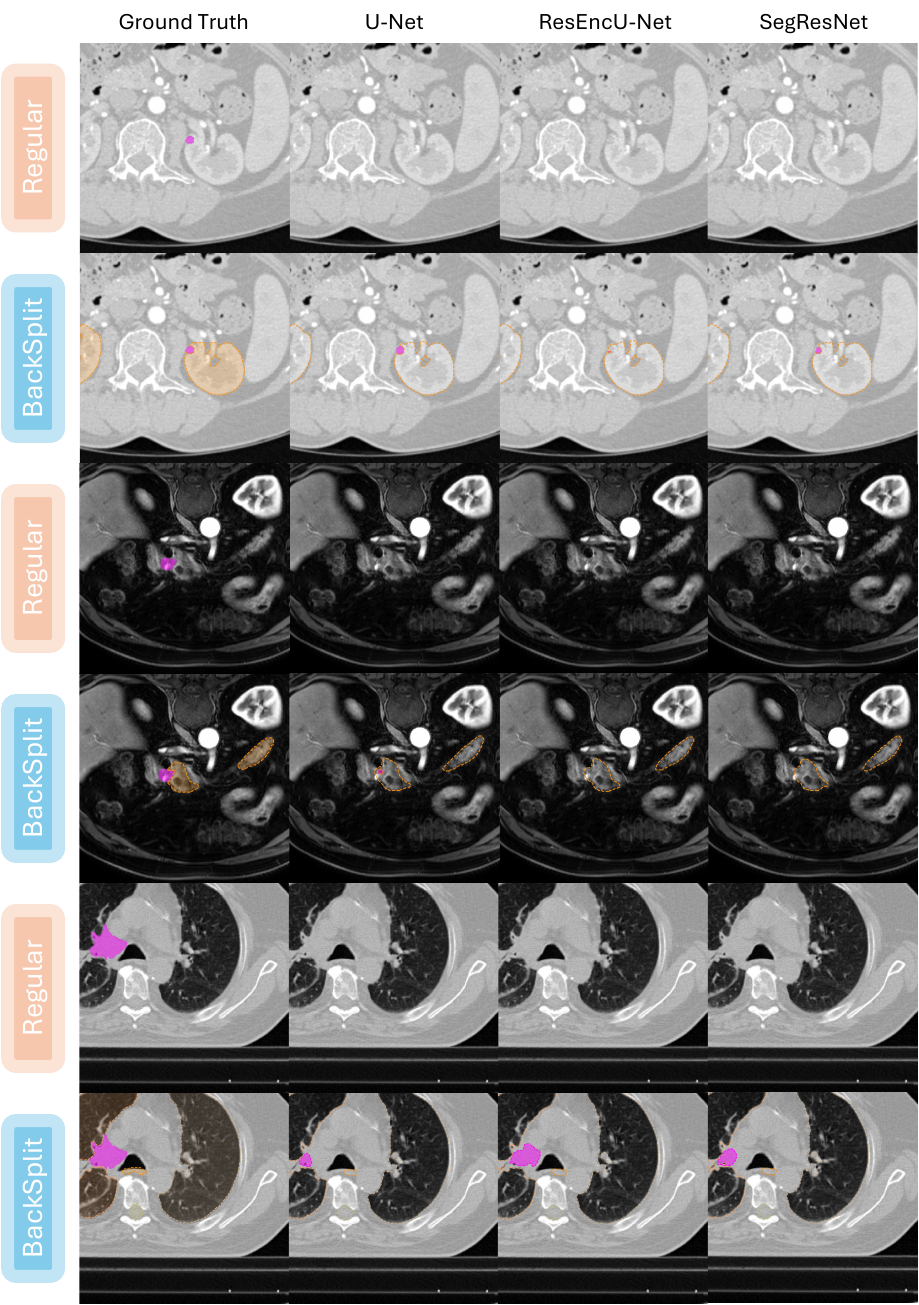}
   \caption{Qualitative comparison on three datasets: KiTS23 (top block), PANTHER-MR (middle block), and NSCLC-Radiomics (bottom block)—across three architectures (U-Net, ResEncU-Net, SegResNet). Each row pair shows regular training (top) and {\methodology} (bottom).
Regular training frequently misses small or low-contrast lesions, whereas {\methodology}, with auxiliary anatomical structures, produces more complete and precise lesion masks. Lesions are shown in pink, while auxiliary support structures predicted under {\methodology}  are shown in yellow/orange hues, highlighting the additional contextual cues that lead to improved segmentation quality.}
   \label{fig:qual_perf1}
\end{figure*}

\begin{figure*}[t!]
  \centering
   \includegraphics[width=0.65\linewidth]{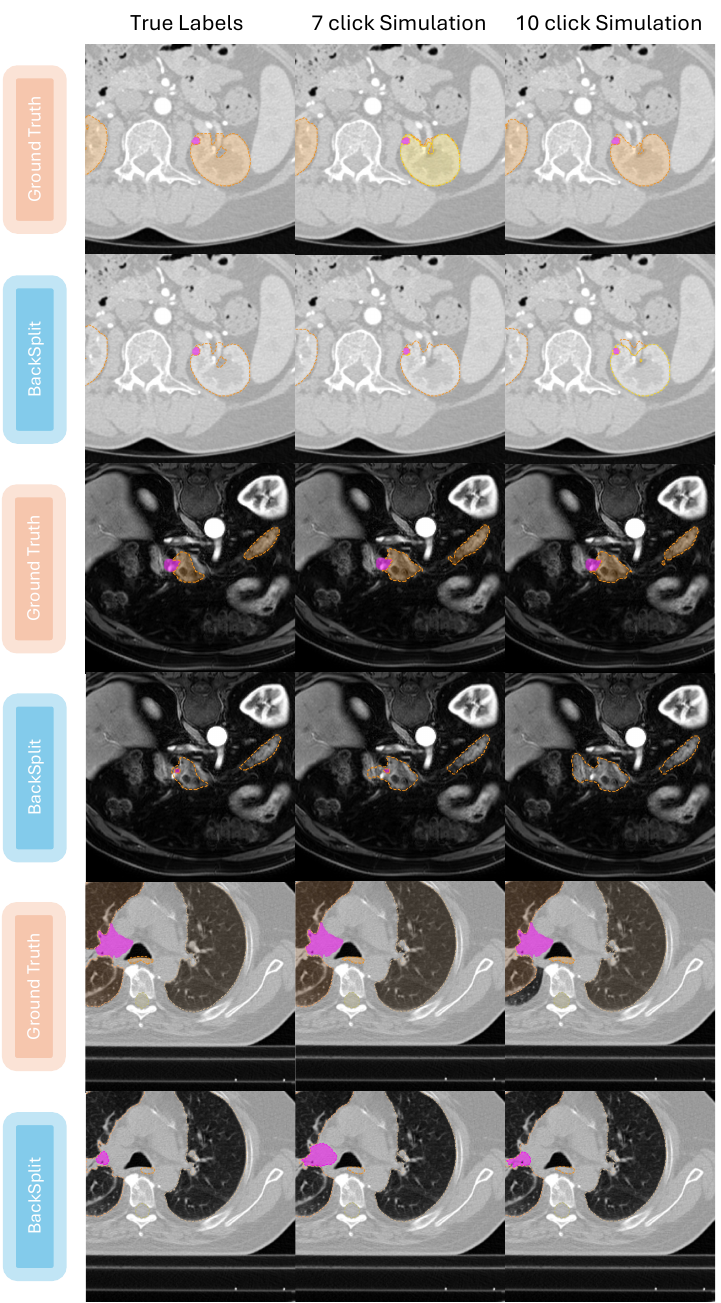}
   \caption{Qualitative comparison of {\methodology} using true auxiliary labels versus noisy auxiliary structures generated by nnInteractive with 7 and 10 positive-click simulations. Rows correspond to three datasets: KiTS23, PANTHER-MR, and NSCLC-Radiomics—and each row pair shows Ground Truth auxiliary labels (top) and {\methodology} predictions (bottom). U-Net backbone is used for all experiments. Lesions are shown in pink, while auxiliary support structures are shown in yellow/orange hues. Despite substantial noise in the interactively generated auxiliary masks, {\methodology} maintains strong lesion segmentation quality, demonstrating robustness to imperfect supervision.}
   \label{fig:qual_perf2}
\end{figure*}

In this section, we present qualitative comparisons illustrating the improvements achieved by {\methodology}. In \cref{fig:qual_perf1}, we compare predictions from standard training and {\methodology} across multiple architectures, highlighting clearer lesion boundaries under {\methodology}. In \cref{fig:qual_perf2}, we show qualitative results using auxiliary labels derived from interactive segmentation. Despite the noisier supervision, {\methodology} continues to produce stable and accurate lesion segmentations, demonstrating its robustness even under imperfect auxiliary annotations.

\end{document}